\documentclass[10pt,letterpaper]{article}
\usepackage[top=0.85in,left=1.0in,footskip=0.75in,marginparwidth=2in]{geometry}
\usepackage{mathrsfs}
\usepackage{amsthm}
\usepackage{amsmath,amssymb,amsfonts}
\usepackage{bm}
\usepackage[utf8]{inputenc}
\usepackage{cite}
\usepackage{nameref,hyperref}
\usepackage{microtype}
\DisableLigatures[f]{encoding = *, family = * }

\raggedright
\usepackage{changepage}
\usepackage[aboveskip=1pt,labelfont=bf,labelsep=period,singlelinecheck=off]{caption}

\usepackage{color}
\definecolor{Gray}{gray}{.25}
\usepackage{graphicx}
\usepackage{sidecap}

\usepackage{wrapfig}
\usepackage[pscoord]{eso-pic}
\usepackage[fulladjust]{marginnote}
\usepackage{graphicx}
\usepackage[section]{algorithm}
\usepackage{algorithmic}
\usepackage{threeparttable}
\usepackage{bm}
\usepackage{color}
\usepackage{ragged2e}
\usepackage{subfigure}
\usepackage{mathrsfs}
\usepackage{enumitem}
\usepackage{array}
\usepackage{url}
\include{multicol}
\newcommand{\PreserveBackslash}[1]{\let\temp=\\#1\let\\=\temp}
\newcolumntype{C}[1]{>{\PreserveBackslash\centering}p{#1}}
\newcolumntype{R}[1]{>{\PreserveBackslash\raggedleft}p{#1}}
\newcolumntype{L}[1]{>{\PreserveBackslash\raggedright}p{#1}}

\renewcommand{\raggedright}{\leftskip=0pt \rightskip=0pt plus 0cm}

\makeatletter
\renewcommand{\@biblabel}[1]{\quad#1.}
\makeatother

\usepackage{lastpage,fancyhdr,graphicx}
\usepackage{epstopdf}
\fancyheadoffset[L]{2.25in}
\fancyfootoffset[L]{2.25in}

\reversemarginpar

\begin{document}
\vspace*{0.35in}

% title goes here:
\begin{flushleft}
{\Large{\textbf{Dynamic Planning of Bicycle Stations in Dockless Public Bicycle-sharing System Using Gated Graph Neural Network}}}
\newline
% authors go here:
\\
Jianguo~Chen\textsuperscript{1},
Kenli~Li \textsuperscript{1},
Keqin Li\textsuperscript{1,2},
Philip S. Yu\textsuperscript{3},
Zeng Zeng\textsuperscript{4}
\\
\bigskip
$^{1}$ College of Computer Science and Electronic Engineering, Hunan University, Changsha, Hunan, 410082, China.
\\
$^{2}$ Department of Computer Science, State University of New York, New Paltz, NY, 12561, USA.
\\
$^{3}$ Department of Computer Science, University of Illinois at Chicago, Chicago, IL, 60607, USA.
\\
$^{4}$ Institute for Infocomm Research, Agency for Science Technology and Research (A*STAR), 138632, Singapore.
\bigskip
\\
* Correspinding author: Kenli Li (lkl@hnu.edu.cn).

\end{flushleft}

\section*{Abstract}
Benefiting from convenient cycling and flexible parking locations, the Dockless Public Bicycle-sharing (DL-PBS) network becomes increasingly popular in many countries.
However, redundant and low-utility stations waste public urban space and maintenance costs of DL-PBS vendors.
In this paper, we propose a Bicycle Station Dynamic Planning (BSDP) system to dynamically provide the optimal bicycle station layout for the DL-PBS network.
The BSDP system contains four modules: bicycle drop-off location clustering, bicycle-station graph modeling, bicycle-station location prediction, and bicycle-station layout recommendation.
In the bicycle drop-off location clustering module, candidate bicycle stations are clustered from each spatio-temporal subset of the large-scale cycling trajectory records.
In the bicycle-station graph modeling module, a weighted digraph model is built based on the clustering results and  inferior stations with low station revenue and utility are filtered.
Then, graph models across time periods are combined to create a graph sequence model.
In the bicycle-station location prediction module, the GGNN model is used to train the graph sequence data and dynamically predict bicycle stations in the next period.
In the bicycle-station layout recommendation module, the predicted bicycle stations are fine-tuned according to the government urban management plan, which ensures that the recommended station layout is conducive to city management, vendor revenue, and user convenience.
Experiments on actual DL-PBS networks verify the effectiveness, accuracy and feasibility of the proposed BSDP system.

\section{Introduction}
With the advantages of zero carbon emissions and convenient cycling, public bicycles have significant benefits in urban short trips and are widely used as public transportation to solve the first/last mile problem \cite{tii01, b01, chen2019gated}.
In many cities in the world, there are numerous Station/Dock-based Public Bicycle-sharing (SD-PBS) systems that provide citizens with public bicycles \cite{li2015traffic, b10}.
In the SD-PBS system, each bicycle station has multiple fixed docks, which greatly limits the movement of the station and the increase of bicycles, and further prevents users from renting and returning bicycles \cite{b03}.
In current years, Dockless PBS (DL-PBS) systems become increasingly prevalent in many countries \cite{pan2019deep, b05, zhang2017bi}.
DL-PBS providers deploy public bicycles at flexible parking points (drop-off locations) instead of fixed stations, and users can return bicycles at anywhere near their destination.
Benefitting from the dockless parking and low station moving cost, we can deploy stations and dispatch bicycles dynamically according to the actual demands in different periods.

Various practical problems arise during the deployment of DL-PBS networks, such as unreasonable bicycle station layout, inefficient and imprecise bicycle deployment, and difficulty in bicycle maintenance \cite{b03}.
On the one hand, due to low deployment costs and vicious competition from peers, a large number of redundant bicycles are deployed in locations that are not frequently used.
Meanwhile, due to the fierce competition between different providers and the low deployment cost, a large number of redundant bicycle stations are arbitrarily deployed \cite{chen2020citywide}.
Massive cluttered bicycle stations from different providers raise the problems of urban road management and traffic safety, while caused a waste of bicycling resources at low-usage stations.
On the other hand, because users may return their bicycles to any place, bicycle drop-off locations will spread across all corners of the city.
Users face difficulties in finding nearby drop-off locations to rent or return bicycles.
Therefore, bicycle station layout is an important factor affecting the Quality of service (QoS) of PBS, and also affects the bicycle deployment, maintenance, and dispatching.

Numerous efforts have been devoted to the research of PBS networks, such as station layout modeling, bicycle dispatching, and bicycle demand prediction \cite{b06, chen2020citywide}.
In station layout modeling, most existing works focused on the traditional SD-PBS networks, mainly relying on empirical experience and surveys in terms of population distribution and environmental factors \cite{b14, b10}.
The traditional approaches face limitation in the novel DL-PBS networks in two aspects:
(1) the large-scale actual bicycle loan datasets are not fully utilized in station modeling, while the real-time bicycle demand is not considered as well,
and (2) SD-PBS networks hold the characteristics that the fixed stations locations and fixed docks in each station, which increase the cost of station migration and have difficulty in matching the flexible bicycle demands.
Existing bicycle stations of the SD-PBS and DL-PBS networks are usually deployed based on the flow rate and density of visitors, lacking in a scientific and reasonable basis.
Moreover, most of existing methods design the static bicycle station layout at the initial phase.
Owing to the high construction cost, once these stations are built, the location of these stations will not change.

\begin{figure}[!ht]
  \centering
  \includegraphics[width=4.5in]{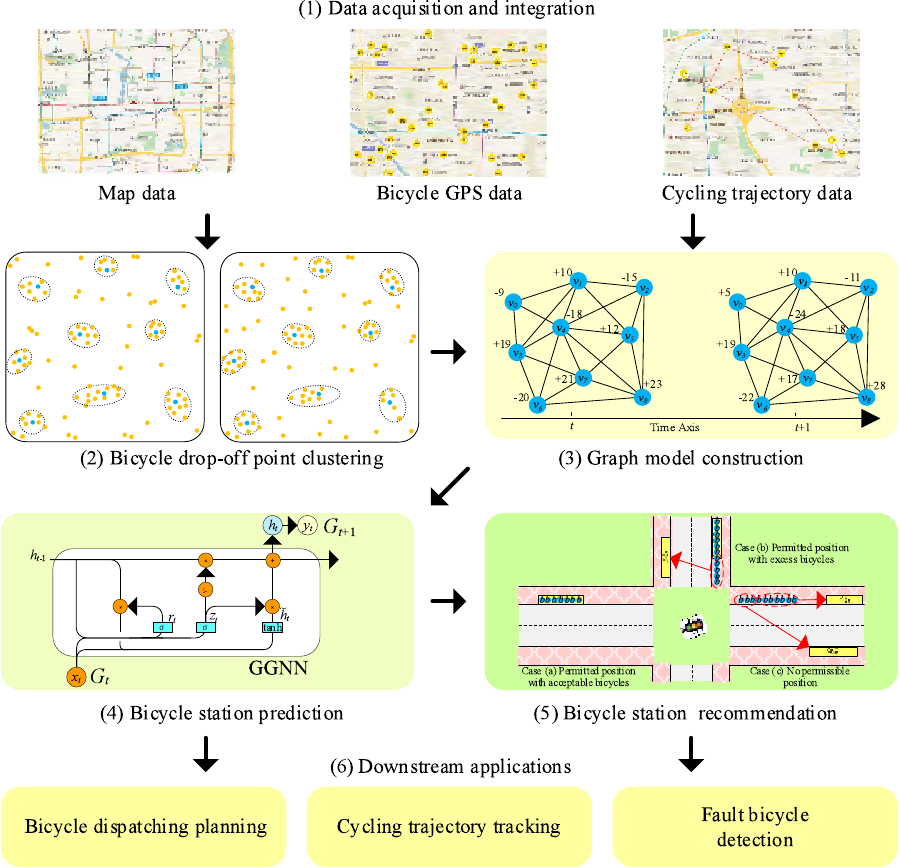}
  \caption{Workflow of the proposed Bicycle Station Dynamic Planning (BSDP) system.
  (1) Collect and integrate large-scale bicycle GPS datasets, cycling trajectory datasets, and map data from the actual DL-PBS network.
  (2) Use a bicycle drop-off location clustering method to detect candidate bicycle stations.
  (3) Create a weighted digraph model based on the candidate bicycle stations after filtering inferior stations, and build a graph sequence model by linking graph models across time periods.
  (4) Use the GGNN model to train the graph sequence data and dynamically predict the bicycle stations in the next period.
  (5) Fine-tune the predicted bicycle stations according to the government's urban management plan to ensure their legitimacy and maximize revenue.
  (6) The bicycle station layout can be used for downstream applications, such as bicycle dispatching planning, cycling trajectory tracking, and fault bicycle detection.}
  \label{fig01}
\end{figure}

In this paper, we focus on the bicycle station layout of the DL-PBS network and propose a Bicycle Station Dynamic Planning (BSDP) system.
The workflow of the BSDP system is illustrated in Fig. \ref{fig01}.
The system can dynamically predict the location of bicycle stations and the number of bicycles needed at each station, providing accurate planning for bicycle station deployment and bicycle dispatching.
Our contributions in this paper can be summarized as follows.
\begin{itemize}
\item We collect large-scale cycling trajectory records from the DL-PBS network and propose a bicycle drop-off location clustering method, in which dense bicycle drop-off locations are clustered as candidate bicycle stations.
\item Based on the clustering results, we construct a bicycle station graph model for each spatio-temporal subset, and remove the inferior stations with low station revenue and utility.
    In addition, graph models across time periods are linked as a graph sequence model.
\item We introduce the Gated Graph Neural Network (GGNN) to create a bicycle station prediction method, in which bicycle demand and bicycle stations in the next time period are dynamically predicted based on the historical graph sequence model.
\item We propose a bicycle station layout recommendation method, where the location of predicted stations and the number of public bicycles needed at each station are fine-tuned according to the government management plan.
\end{itemize}

The remainder of the paper is organized as follows.
Section \ref{section2} reviews the related work.
Section \ref{section3} introduces the bicycle drop-off location clustering and graph modeling of the DL-PBS network.
Section \ref{section4} describes the bicycle station dynamic planning system.
Section \ref{section5} provides experimental evaluations.
Finally, Section \ref{section6} concludes this paper with future work and directions.

\section{Related Work}
\label{section2}
Various efforts have been made in researching big data and data mining techniques to build smart cities \cite{zhang2017bi, tii01, zhang2017contention}.
Numerous studies of PBS networks and smart cities attract both fields of industry and academia \cite{b04, b06, chen2018exploiting}.
Focusing on the analysis and prediction of public bicycle demands, Zhang \emph{et al}. used a circular distribution method to obtain the daily peak of the public bicycle trips.
They used a time series model to predict the bicycle demands during rush hours \cite{chen2018exploiting}.
Etienne \emph{et al}. analyzed the use of PBS networks by using model-based count series clustering \cite{b01}.
Cazabet \emph{et al}. studied cycling rules from the perspective of signal processing and data analysis \cite{b11}, in which a cycling periodic model was built by analyzing the characteristics of time, space, and riders.
However, most existing models are built on external factors such as population density and travel probability, rather than the laws of cycling trajectory records themselves.

There are numerous approaches focused on location prediction \cite{jia2016location} in temporal-spatial environments.
In \cite{jia2016location}, Jia \emph{et al}. proposed a temporal-spatial Bayesian model to predict user's location based on his influential friends' locations.
In \cite{ying2014mining}, Ying \emph{et al}. introduced a Geographic-Temporal-Semantic (GTS) model and proposed a GTS-based location prediction method.
They collected the trajectories of users and calculated the similarity of the movement and trajectories between users.
Focusing on the layout of public bicycle stations, various schemes were carried out in \cite{b03, b14}.
Ma \emph{et al}. proposed a hierarchical public bicycle dispatching strategy for dynamic demand \cite{b03}.
In \cite{b13}, Deng \emph{et al}. discussed the layout optimization of public bicycle stations based on the AHP method.
Jiang \emph {et al}. analyzed the GPS trajectory of urban bicycles and detected $K$ primary corridors on the road network \cite{b14}.
The scale of bicycle stations includes the number of bicycles, the grade of the station, and the scope of bicycle-sharing services.
However, most of the existing studies focus on the traditional SD-PBS network, and they design the static bicycle station layout at the initial phase.
Owing to the high construction cost, once these stations are built, the location of these stations will not change.
Different from the existing research, we focus on the dynamic layout of the bicycle stations for the DL-PBS network, and update the bicycle stations between different time periods according to the actual bicycle usage demands.

Graph Neural Network (GNN) is an effective deep learning model used in graph or network applications \cite{chen2019gated, gnn05, gnn01, gnn06}.
In \cite{gnn03}, Khodayar \emph{et al}. proposed a GNN model and applied it to wind speed prediction.
Levie \emph{et al}. introduced a spectral-domain convolutional architecture of DL on graph models \cite{gnn04}.
Lin \emph{et al}. used the GNN method to predict the hour-level demands of bicycle stations in the SD-PBS network.
They used the Long Short-Term Memory (LSTM) neural network to capture the temporal dependency in bicycle-sharing demand sequences \cite{b15}.
In \cite{b16}, Gast \emph{et al}. introduced a generalized regression neural network to predict the public bicycle demands of SD-PBS.
Li \emph{et al}. modified the GNN model and proposed a Gated Graph Neural Network (GGNN) to achieve a flexible and broadly output sequence \cite{gnn08}.
In this work, we use the GGNN model to train the bicycle station graph sequence of the DL-PBS network and predict the location of bicycle stations and their bicycle-sharing demands.

\section{DL-PBS Network Clustering and Graph Modeling}
\label{section3}
In this section, we will describe the DL-PBS network and highlight its characteristics with practical issues that arise during the deployment.
Then, based on the cycling trajectory records, we establish a weighted digraph model for the DL-PBS network, and calculate the update of bicycle stations to create a graph sequence model.

\subsection{Dockless Public Bicycle-sharing (DL-BPS) Network}
\begin{figure}[!ht]
 \centering
 \subfigure[Dockless stations]{\includegraphics[width=1.5in]{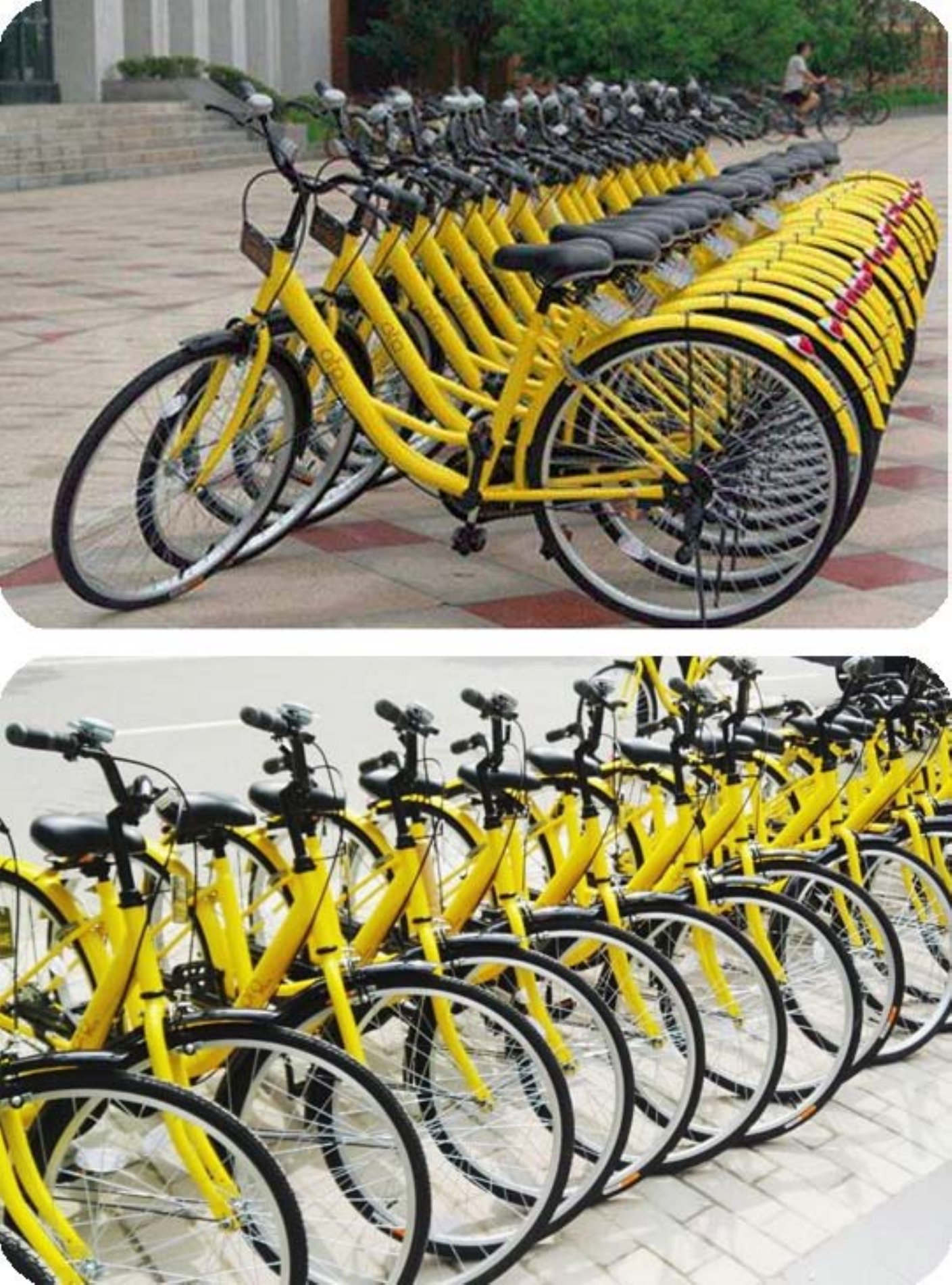}}
 \subfigure[Scanning QR code]{\includegraphics[width=1.5in]{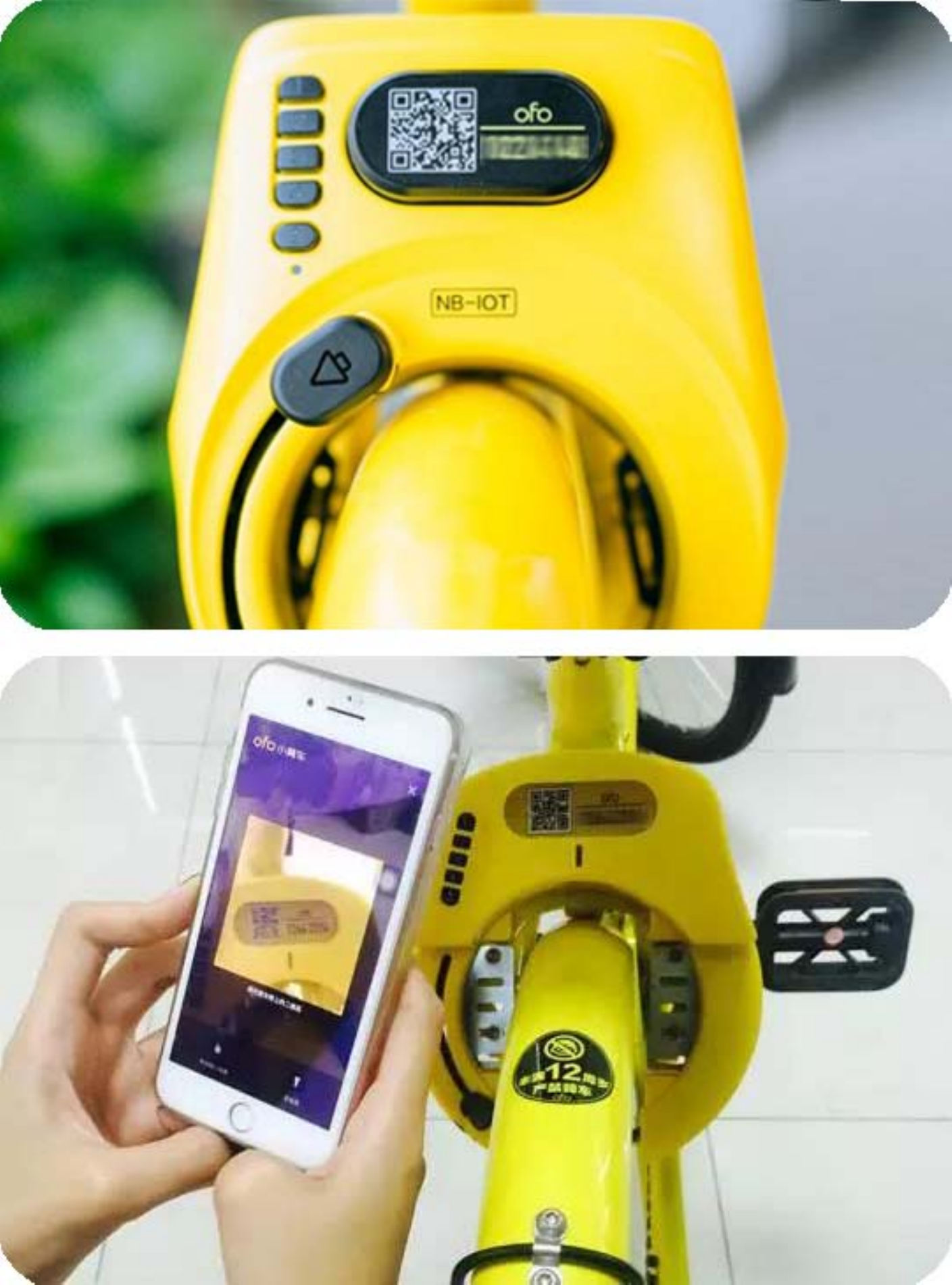}}
  \subfigure[Mobile App]{\includegraphics[width=1.07in]{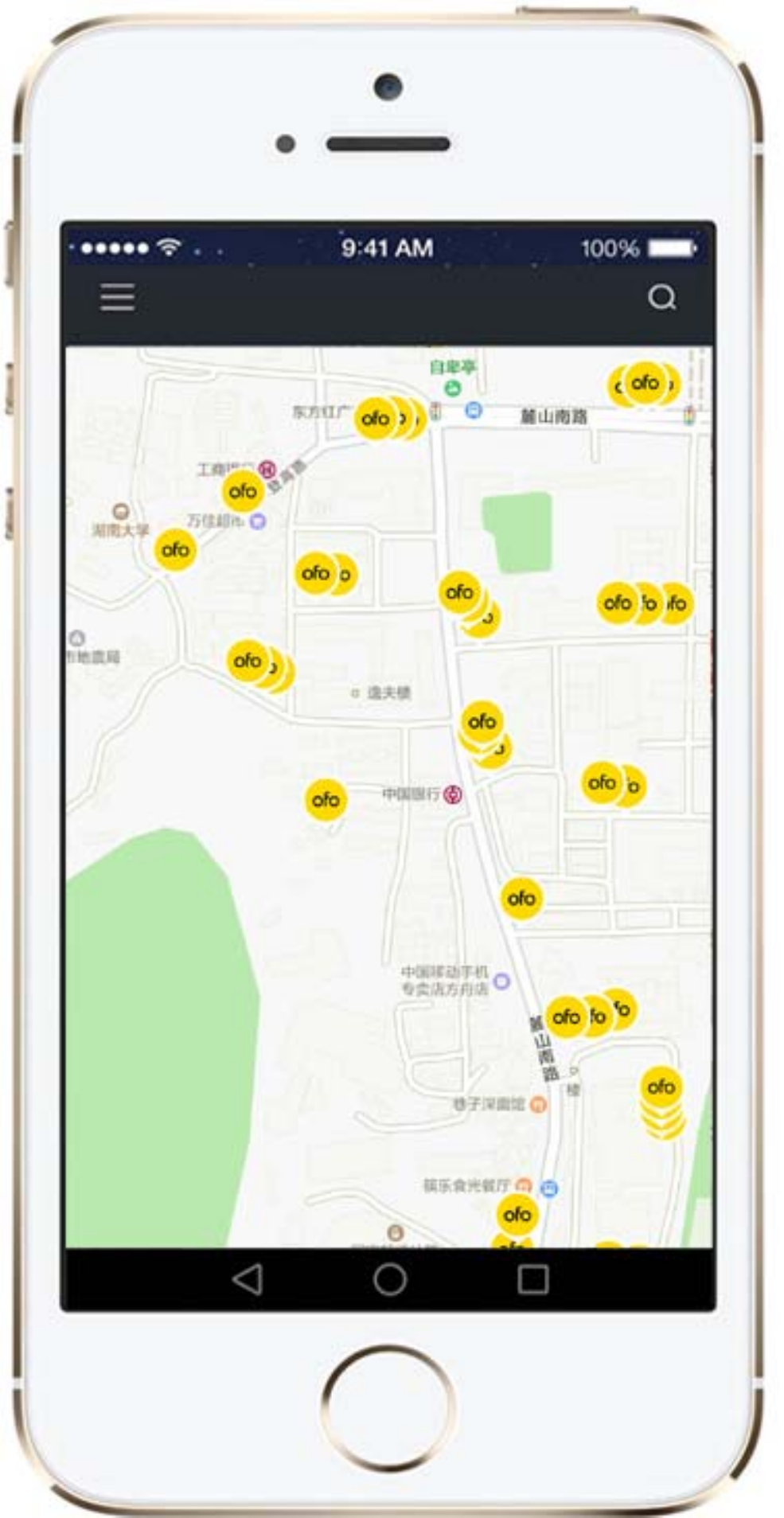}}
 \caption{Principal components of a DL-PBS network, including public bicycles, dockless bicycle stations, a GPS module, a QR-code-based locking module, and a mobile application.}
 \label{fig02}
\end{figure}

As a new generation of PBS systems, the DL-PBS network provides personalized and convenient services during bicycle rental and return \cite{pan2019deep}.
The principal components of a DL-PBS network include public bicycles, dockless stations, Global Positioning System (GPS), a Quick Response (QR) code-based locking module, and a mobile application, as shown in Fig. \ref{fig02}.

\begin{itemize}
\item Public bicycles.
    Public bicycles have a distinctive look to be quickly identified by users.
    Each bicycle is equipped with a GPS to record its position in real time and a QR code-based locking module.
\item Dockless bicycle stations (also called drop-off locations or parking points).
    The DL-PBS providers deploy bicycles at any permitted public parking areas, such as the curbside, the entrances of parks, communities, and shopping malls.
    Each dense bicycle parking point is termed as a temporary dockless bicycle station, as shown in Fig. \ref{fig02}(a).
\item QR code.
    Each bicycle has a unique QR code that can be scanned via a mobile App to unlock the bicycle and pay the rent, as shown in Fig. \ref{fig02}(b).
\item Mobile application (APP).
    Mobile APP is an important component of the DL-PBS network.
     It provides functions including bicycle GPS locating, QR code scanning and unlocking, payment, and cycling trajectory tracking, as shown in Fig. \ref{fig02}(c).
\end{itemize}

\subsection{Bicycle Drop-off Location Clustering}
\begin{figure}[!ht]
  \centering
 \subfigure[Cycling trajectory records]{\includegraphics[width=2.7in]{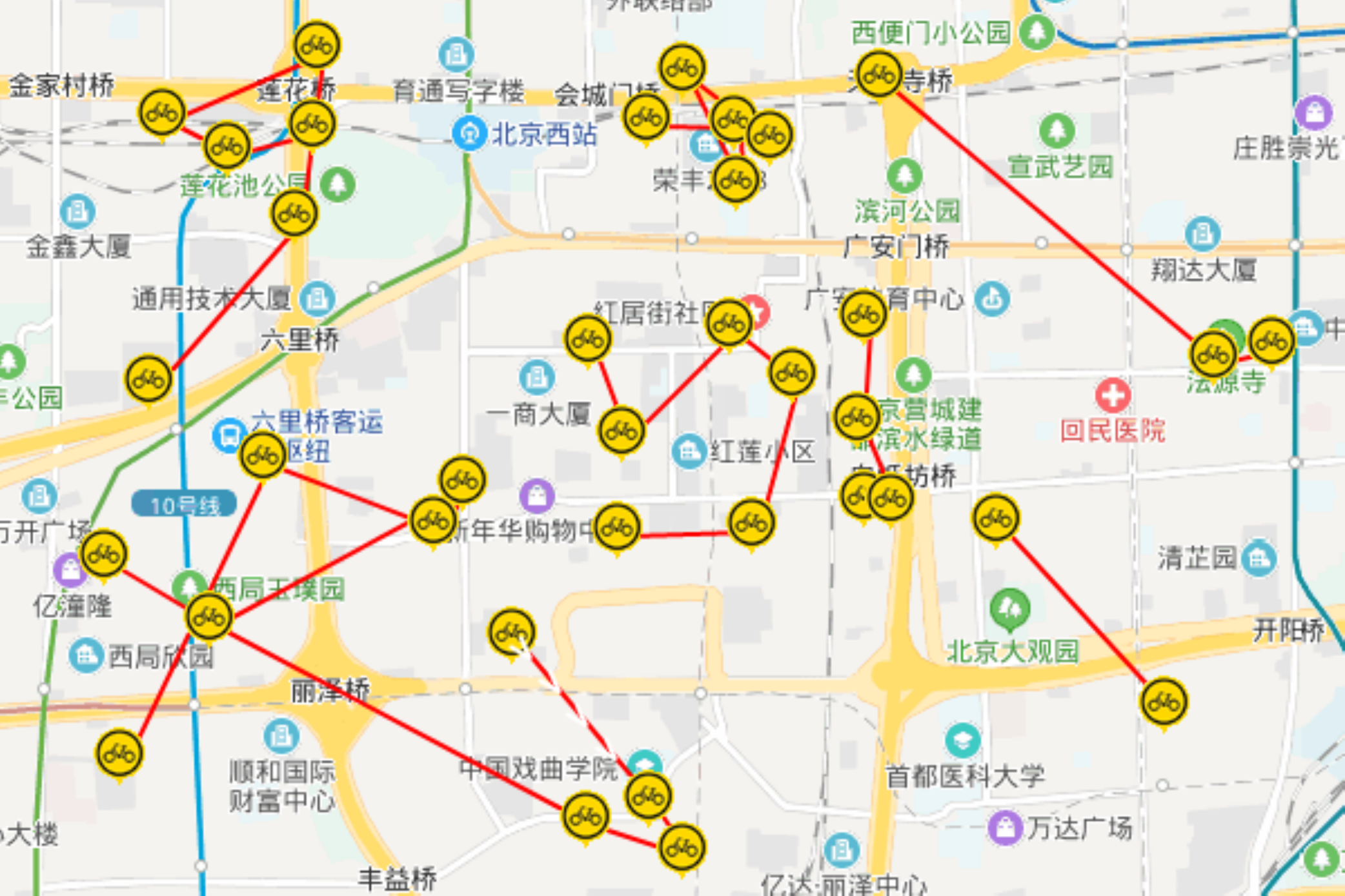}}
 \subfigure[Bicycle GPS records]{\includegraphics[width=2.7in]{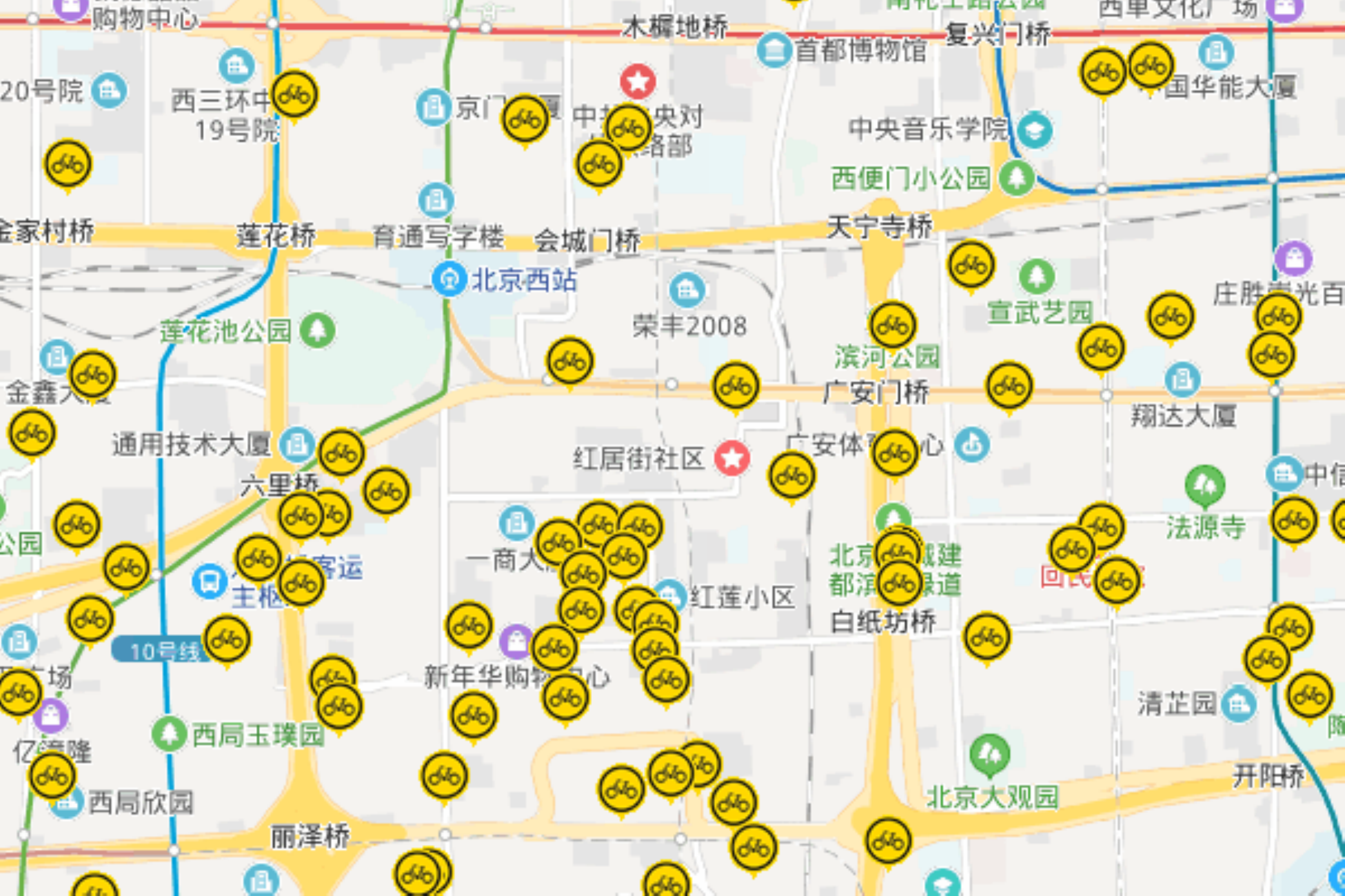}}
\caption{Examples of cycling trajectory and GPS records. (a) is the cycling trajectory records of bicycles, where each yellow icon is a bicycle GPS location, and the red directed line is the cycling route. (b) is the GPS records of stationary bicycles.}
\label{fig03}
\end{figure}

We collect large-scale historical bicycle GPS datasets and cycling trajectory records from DL-PBS networks deployed in different cities and administrative regions.
Bicycle GPS records are routinely collected from all stationary bicycles.
Each bicycle GPS record contains the bike ID, longitude, latitude, and the timestamp of collection.
Cycling trajectory records are saved from the users' cycling behaviors.
Each cycling trajectory record contains user ID, bike ID, the timestamp of pick up (rent) and drop off (return), longitude and latitude of rent and return positions, cycling distance and cost.
Considering that the layout of bicycle stations is usually suitable for a certain city or administrative region, we need to build a bicycle station graph model for each city or administrative region at each time period.
Therefore, we divide the cycling trajectory dataset into multiple spatial subsets based on administrative divisions, and further divide each spatial subset into multiple temporal subsets by time periods.
Examples of cycling trajectory and GPS records are given in Fig. \ref{fig03} and Table \ref{table01}.

\begin{table}[!ht]
\renewcommand{\arraystretch}{1.3}
\caption{Examples of cycling trajectory records.}
\label{table01}
\centering
\tabcolsep1pt
\begin{tabular}{C{0.25in} C{0.5in} C{1.0in} C{0.6in} C{0.6in} C{1.0in} C{0.6in} C{0.6in}}
\hline
     &             &\multicolumn{3}{c}{Departure Info.}    &\multicolumn{3}{c}{Arrival Info.}\\
\hline
User &	Bicycle & Time stamp &	Latitude &	Longitude &	Time stamp & Latitude &	Longitude \\
\hline
01	& e1xx4 &	2018/10/25 10:20:22 & 39.914548	& 116.440848 &	2018/10/25 10:48:13 & 39.900323 & 116.484110\\
02	& e1xx9 &	2018/10/25 09:11:19	& 39.914326	& 116.482170 &	2018/10/25 09:43:27 & 39.899604 & 116.425325\\
03	& elxx3 &	2018/10/25 19:15:10	& 39.899604	& 116.425325 &	2018/10/25 19:44:23	& 39.890705	& 116.483715\\
    & \dots & \dots & \dots & \dots & \dots & \dots & \dots\\
\hline
\end{tabular}
\end{table}

The bicycle drop-off locations collected from bicycle GPS datasets have the characteristics of varying density distribution (VDD), equilibrium distribution (ED), and multiple domain-density maximums (MDDM).
Specifically, bicycles are densely distributed in some areas (i.e., urban areas), while sparsely distributed in other areas (i.e., suburban areas), which is in line with the VDD characteristic.
In addition, in the initial deploy state, the bicycles at each drop-off location are parked neatly at the same interval, so each bicycle at the drop-off location has an ED.
Moreover, it is easy to know that the bicycle position may have multiple domain density maximums during use and return.
The Domain Adaptive Density Clustering Algorithm (DADC) algorithm proposed in our previous work \cite{chen2019} can obtain more reasonable clustering results on data with VDD, ED, and MDDM characteristics.
Therefore, we introduce the DADC algorithm for bicycle drop-off location clustering.

Given a set of cycling trajectory records in a temporal subset, we extract all GPS information of pick up and drop off from each record.
Assume that there are $M$ records in the dataset $X$, hence $2M$ positions are extracted.
Considering that the same bicycle may be used many times during the current period, the position will be extracted repeatedly.
In the rest of this paper, we perform bicycle drop-off location clustering and graph modeling for each temporal subset $X$.
We filter the position information of the same bicycle ID with the same latitude and longitude.
We get the dataset of bicycle positions in the current period $X=\{x_{1}, \dots, x_{N}\}$, and the number of positions $N$ satisfies $M \leq N \leq 2M$.
For each data point $x_{i}$ in $X$, we define the local density $\rho_{i}$ of $x_{i}$ as:
\begin{equation}
\label{eq01}
\rho_{i} = \sum_{x_{j} \in N(x_{i})}{\chi (d_{ij} - d_{c})},
\end{equation}
where $N(x_{i})$ is the neighbors of $x_{i}$, $d_{c}$ is a cutoff distance, and $\chi (d_{ij} - d_{c}) = 1$, if $d_{ij} < d_{c}$; otherwise, $\chi (d_{ij} - d_{c}) = 0$.
Therefore, $\rho_{i}$ is equal to the number of points closer than $d_{c}$ to $x_{i}$.
In addition, we calculate the delta distance $\delta_{i}$ of $x_{i}$ by computing the shortest distance between $x_{i}$ and any other data points with a higher density:
\begin{equation}
\label{eq02}
\delta_{i} = \min_{j:\rho_{j} > \rho_{i}}{d_{ij}}.
\end{equation}
For the highest density point, $\delta_{i} = \max_{x_{j} \in X}{(d_{ij})}$.

\begin{figure}[!ht]
  \centering
 \subfigure[Bicycle drop-off location clustering]{\includegraphics[width=2.7in, height=1.93in]{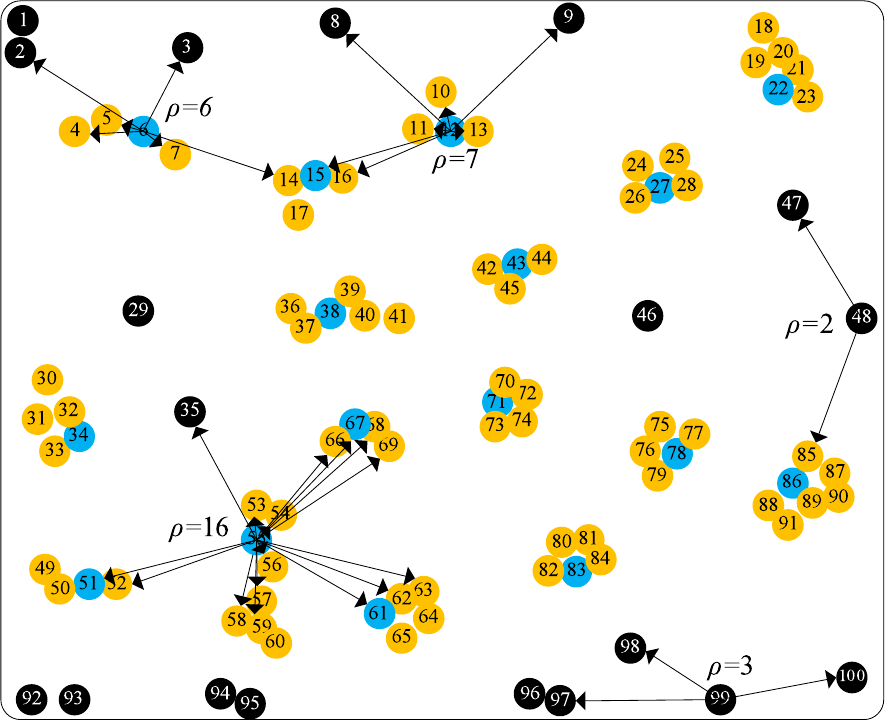}}
 \subfigure[Clustering decision graph]{\includegraphics[width=2.7in, height=1.93in]{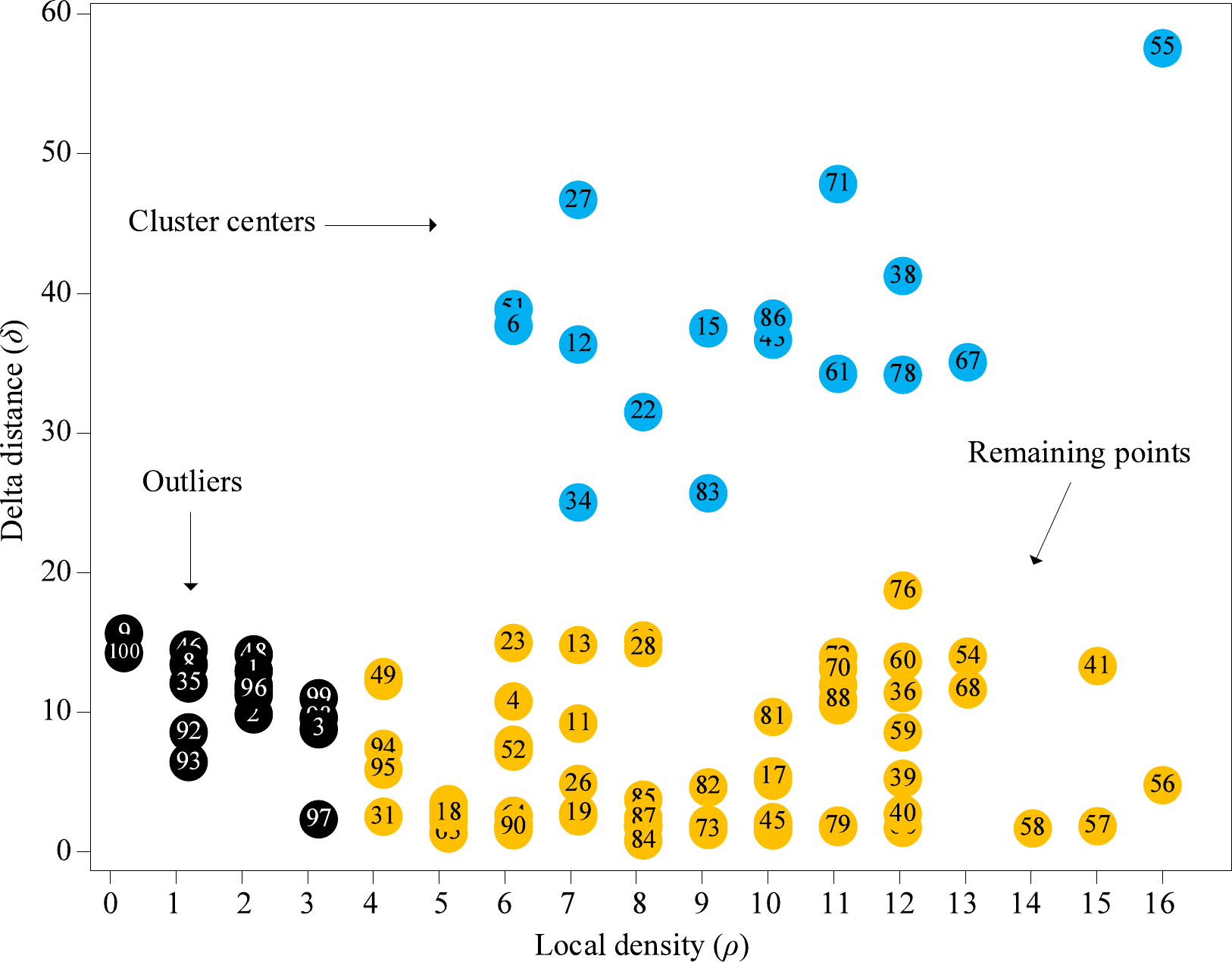}}
  \caption{Example of DADC-based bicycle drop-off location clustering. (a) For each data point, we calculate the local density and the delta distance. (b) We draw a clustering decision graph based on local density and delta distance to detect cluster centers, outliers, and remaining points.}
  \label{fig04}
\end{figure}

After calculating the local density $\rho$ and the delta distance $\delta$, we can draw a clustering decision graph based on $\rho$ and $\delta$, where $\rho$ is the x-axis and $\delta$ is the y-axis.
Then we observe the distribution of these points from the graph.
Data points with a high $\rho$ ($\rho_{i} > \theta_{\rho}$) and a high $\delta$ ($\delta_{i} > \theta_{\delta}$) are considered as cluster centers, while points with a low $\rho$ and a high $\delta$ are considered as outliers.
In practical applications, the values of $\theta_{\rho}$ and $\theta_{\delta}$ are manually set based on experience.
In this work, the effective thresholds are set as $\theta_{\rho} = \frac{1}{3}\rho_{max}$ and $\theta_{\delta} = \frac{1}{3}\delta_{max}$.
After finding the cluster centers, each remaining point is assigned to the same cluster as its nearest neighbor with a higher density.
An example of the DADC-based bicycle drop-off location clustering process is illustrated in Fig. \ref{fig04}.

The detailed steps of DADC-based bicycle drop-off location clustering are presented in Algorithm \ref{alg1}.
The process of the bicycle drop-off location clustering includes sub-processes of cluster center detection and remaining point assignment.
Assuming that the number of data points in $X$ is equal to $N$, the computational complexity of Algorithm \ref{alg1} is $O(N)$.

\begin{algorithm}[!ht]
\caption{DADC-based bicycle drop-off location clustering}
\label{alg1}
\begin{algorithmic}[1]
\REQUIRE ~\\
    $X$: A temporal subset of cycling trajectory records;\\
    $\theta_{\rho}$: The local density threshold for the data points in $X$;\\
    $\theta_{\delta}$: The delta distance threshold for the data points in $X$;\\
\ENSURE ~\\
    $C$: The clustered bicycle stations.
\STATE extract all bicycle locations of rent and return from $X_{raw}$;
\STATE filter duplicate positions and obtain a set of bicycle locations $X$;
\FOR {each data point $x_{i}$ in $X$}
\STATE calculate the local density $\rho_{i}$ using Eq. (\ref{eq01});
\STATE calculate the delta distance $\delta_{i}$ using Eq. (\ref{eq02});
\IF  {$\rho_{i} > \theta_{\rho}$ and $\delta_{i} > \theta_{\delta}$}
\STATE mark $x_{i}$ as a cluster center $C \leftarrow x_{i}$;
\ELSE
\STATE mark $x_{i}$ as a remaining point $\Lambda \leftarrow x_{i}$;
\ENDIF
\ENDFOR
\FOR {each remaining point $x_{i}$ in $\Lambda$}
\STATE assign $x_{i}$ to the nearest cluster $c$;
\ENDFOR
\RETURN $C$.
\end{algorithmic}
\end{algorithm}

\subsection{Bicycle Station Graph Model}
\begin{figure}[!ht]
\centering
 \subfigure[Actual DL-PBS network]{\includegraphics[width=2.2in]{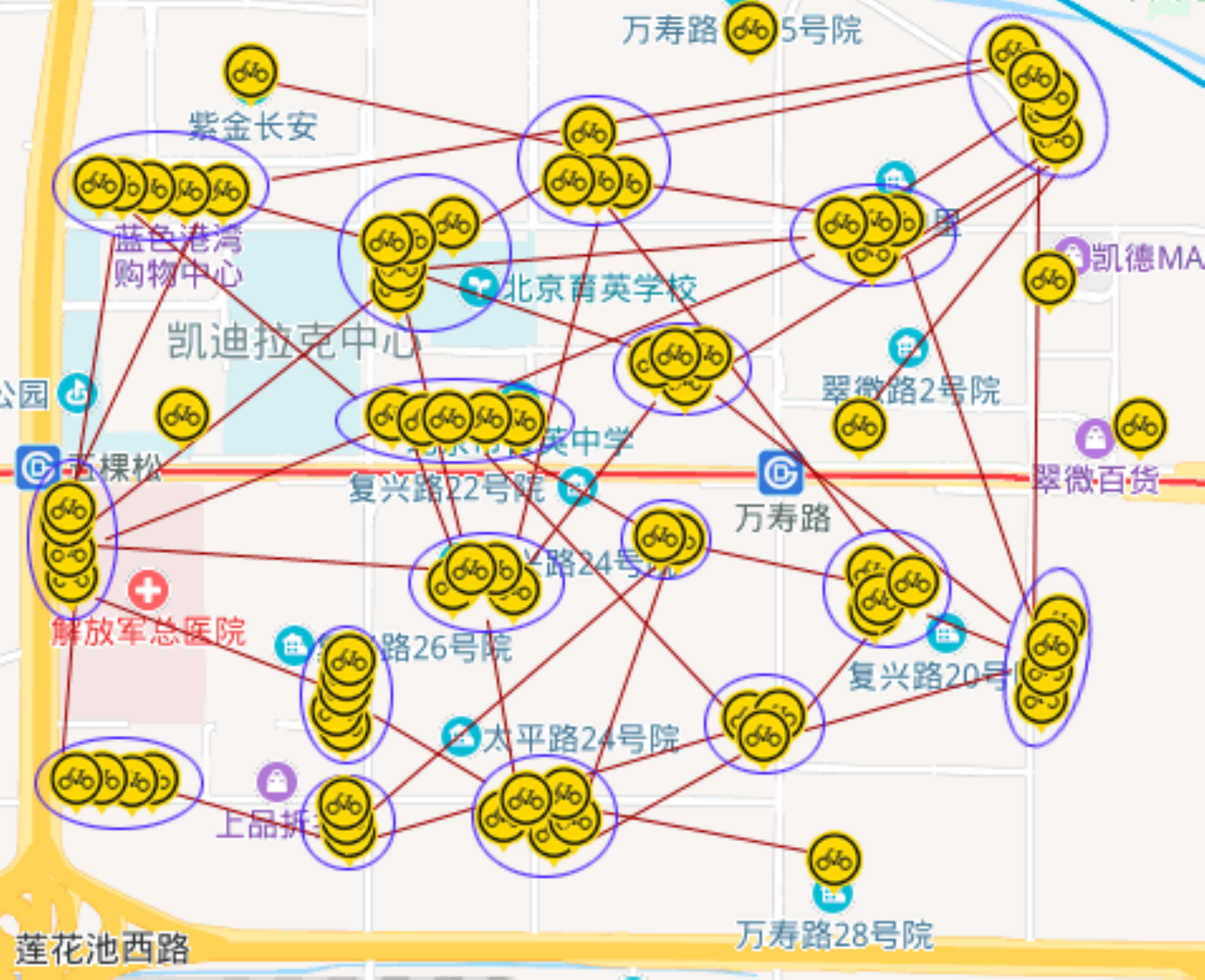}}
 \hspace{0.1in}
 \subfigure[Weighted digraph model]{\includegraphics[width=2.2in]{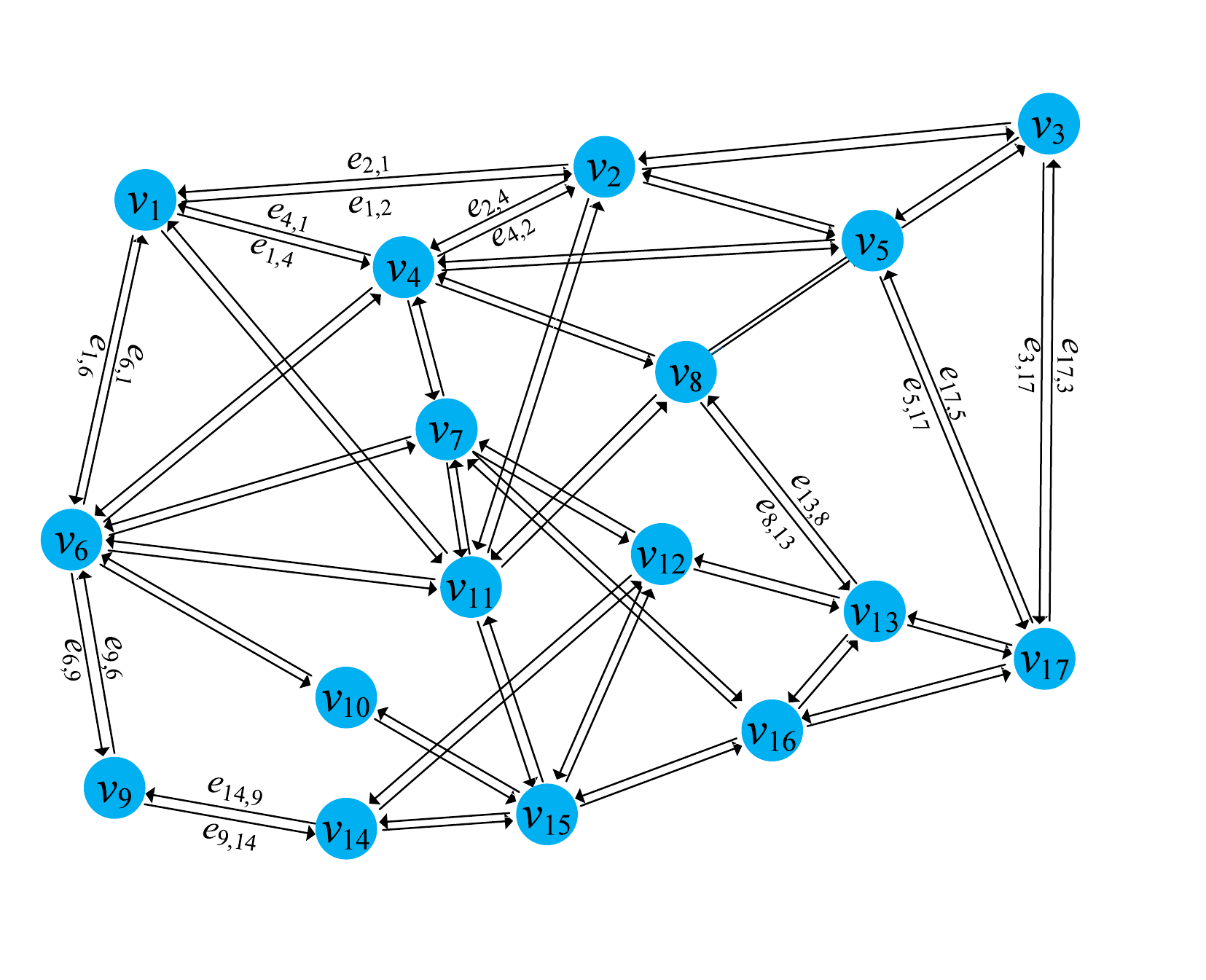}}
 \caption{Graph modeling of DL-PBS bicycle stations. The clusters of bicycle drop-off locations are treated as candidate bicycle stations and used as the vertices of the station graph model. In addition, the cycling records between stations are collected as the corresponding directed edges between vertices.}
 \label{fig05}
\end{figure}

Based on the bicycle drop-off location clustering results, we obtain a series of candidate bicycle stations.
We build a weighted digraph model for these stations, where the stations are considered as vertices, and the cycling records between them are collected as the corresponding directed edges.
In addition, we detect inferior stations with low station revenue and utility and remove them from the graph to obtain a high-quality bicycle station graph.
An example of the bicycle station graph modeling process is shown in Fig. \ref{fig05}.

(1) Construct a weighted digraph for bicycle stations.

A DL-PBS network can be intuitively modeled as a weighted digraph model, where vertices are bicycle stations and edges are cycling trajectories between them.
Let $G = (V, ~E, ~D, ~W)$ be a weighted digraph model of the DL-PBS network in a temporal subset, where $V$ is the set of bicycle stations, $E$ is the set of cycling paths between stations, $D$ is the actual distance between stations, and $W$ is the number of cycling records between stations.
We calculate the value of the actual distance $D = \{ \dots, d_{ij}, \dots \}$, where each element $d_{ij}$ represents the distance between stations $v_{i}$ and $v_{j}$, and $d_{ij} = d_{ji}$.
Considering that each bicycle station is a cluster of public bicycles, we find the cluster center of each station.

Let $\psi_{i}$ and $\varphi_{i}$ be the latitude and longitude of station $v_{i}$, we can use the Haversine method to calculate the distance between stations $v_{i} (\varphi_{i}, \psi_{i})$ and $v_{j} (\varphi_{j}, \psi_{j})$:
\begin{equation}
\label{eq03}
%\begin{aligned}
d_{ij} = 2 \times R \times \sin{\sqrt{H\biggl(\frac{d}{R}\biggr)}}
       = 2R \times \sin{\sqrt{H(|\psi_{i}-\psi_{j}|)+\cos(\psi_{i})\cos(\psi_{j})H(|\varphi_{i}-\varphi_{j}|)}},
%\end{aligned}
\end{equation}
where $R$ is the radius of the earth, usually set to 6371.0 km, and the Haversine function $H(\theta)$ is defined as:
\begin{equation}
\label{eq04}
H(\theta) = \sin^{2}\biggl(\frac{\theta}{2}\biggr) = \frac{1}{2}(1- \cos(\theta)).
\end{equation}

For each directed edge $e_{ij}$, we count the number of cycling records starting from station $v_{i}$ and arriving at $v_{j}$, and treat it as the weight $w_{ij}$ of the edge.
Note that $w_{ij} \neq w_{ji}$.
In this way, we obtain the set of weights $W$ of all directed edges $E$ and create a weighted digraph model $G = (V, ~E, ~D, ~W)$.

(2) Remove inferior bicycle stations.

As mentioned above, due to low deployment costs and vicious competition from peers, a large number of redundant bicycles are deployed in locations that are not frequently used.
We need to detect and remove these station from the bicycle station graph to maximize the benefits and utility of the DL-PBS network.

\textbf{Definition 1}. Station revenue.
The revenue of a bicycle station is the sum of the cycling costs of all bicycles rented (departing) from the station.
The revenue of each station $v_{i}$ is calculated as:
\begin{equation}
\label{eq05}
P_{i} = \sum\limits_{w_{ij} \in W} {(w_{ij}\times d_{ij} \times \alpha)},
\end{equation}
where $\alpha$ is the cycling cost per unit distance, and $(d_{ij} \times \alpha)$ is the cycling cost of a bicycle from $v_{i}$ to $v_{j}$.

\textbf{Definition 2}. Station utility.
The throughput of a bicycle station refers to the number of bicycles leaving or arriving at this station.
The utility of a station is the ratio of the station's throughput to the entire graph's throughput.
The throughput of a station $v_{i}$ is defined as:
\begin{equation}
\label{eq06}
TP_{i} = \sum\limits_{w_{ij},w_{j'i} \in W}{\left(w_{ij} + w_{j'i}\right)},
\end{equation}
where $w_{ij}$ is the weight of the directed edges leaving $v_{i}$ to any station $v_{j}$ and $w_{j'i}$ is the weight of the directed edges arriving at $v_{i}$ from any station $v_{j}$.
Let $TP_{G} = \sum_{w_{ij} \in W}{w_{ij}}$ be the throughput of the entire graph, we can calculate the utility of station $v_{i}$ as:
\begin{equation}
\label{eq07}
U_{i} = \frac{TP_{i}}{2TP_{G}} = {\sum\limits_{w_{ij},w_{j'i} \in W}{\left(w_{ij} + w_{j'i}\right)}}\left/{2\sum_{w_{ij} \in W}{w_{ij}}}\right..
\end{equation}
The relationship of station utilization and station degrees: utilization is the number of borrowed and returned bicycles at this station, and station degrees refers to the station of the number of access.

\textbf{Definition 3}. Inferior bicycle stations.
A bicycle station $v_{i}$ is regarded as an inferior station if its revenue $P_{i}$ is below the given threshold $\theta_{P}$ and its utility $U_{i}$ is below the given threshold $\theta_{U}$.

Station revenue is positively related to station utility.
Namely, an increase in station utility will bring an increase in station revenue.
In the actual operation of the DL-PBS network, due to different DL-PBS network layouts and densities in different cities, distinct values are set to these thresholds.
Based on the station revenue and utility, we can detect inferior bicycle stations and remove them and their associated edges from the bicycle station graph.

Algorithm \ref{alg2} describes the detailed steps of bicycle station graph modeling.
In Algorithm \ref{alg2}, the process of the bicycle station graph modeling includes sub-processes of graph construction and inferior bicycle stations deletion.
Assuming that the number of bicycle station clusters is equal to $|C|$, that is, the number of vertices of the bicycle station graph $G$, the computational complexity of Algorithm \ref{alg2} is $O(|C|)$ .

\begin{algorithm}[!ht]
\caption{Bicycle station graph modeling}
\label{alg2}
\begin{algorithmic}[1]
\REQUIRE ~\\
    $C$: The clustering results of the bicycle stations;\\
    $X$: A temporal subset of cycling trajectory records;\\
    $\theta_{P}$: The threshold of station revenue;\\
    $\theta_{U}$: The threshold of station utility;\\
\ENSURE ~\\
    $G$: the graph model of the DL-PBS network.
\FOR {each cluster $c_{i}$ in clusters $C$}
\STATE identify the cluster center and create a vertex $v_{i} \leftarrow center(c_{i})$;
\STATE calculate attributes of $\varphi_{i}$, $\psi_{i}$, and $n_{i}$ for $v_{i}$;
\STATE calculate directed edges $E$ and weights $W$ from $C$ and $X$;
\STATE calculate the distances $D$ between vertices using Eq. (\ref{eq03});
\ENDFOR
\STATE build a graph model $G \leftarrow (V,~E,~D,~W)$;
\FOR {each vertex $v_{i}$ in $V$}
\STATE calculate station revenue $P_{i}$ using Eq. (\ref{eq05});
\STATE calculate station utility $U_{i}$ using Eq. (\ref{eq07});
\IF {$P_{i} < \theta_{P}$ and $U_{i} < \theta_{U}$ }
\STATE detect $v_{i}$ as an inferior station and remove $v_{i}$ from $V$;
\STATE remove edges $e_{ij}$ and $e_{ji}$ from $E$;
\ENDIF
\ENDFOR
\RETURN $G$.
\end{algorithmic}
\end{algorithm}

\subsection{Graph Sequence Model of DL-PBS Network}
\begin{figure}[!ht]
\centering
\includegraphics[width=4.5in]{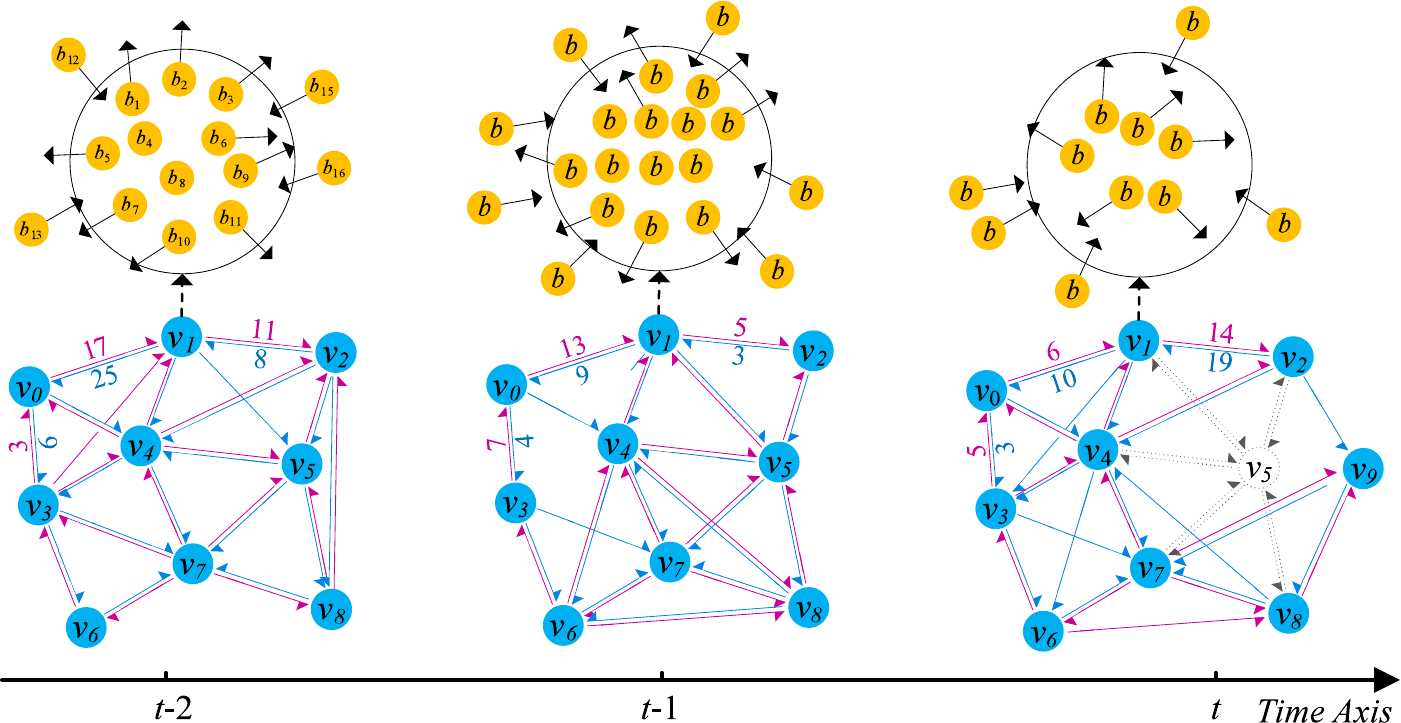}
\caption{Bicycle station updating and graph sequence modeling.}
\label{fig06}
\end{figure}

In the previous section, we established a bicycle station graph for each temporal subset of cycling trajectory records in a city or administrative region.
In this way, we can build a series of graphs between different time periods in the same area.
In this section, we consider updates to bicycle stations across time periods, add a time dimension to these graphs, and build a graph sequence model.
An example of the graph sequence modeling process is illustrated in Fig. \ref{fig06}.

Given a time series $T=\{1, \dots, t\}$, we divide a region's historical cycling trajectory records into a series of temporal subsets $X = {X_{1}, \dots, X_{t}}$, and then build a graph model $G_{t}$ for each subset $X_{t}$.
In this way, we can build a series of graphs and create a graph sequence model $GS = \{G_{1}, \dots, G_{t}\}$.

\section{Bicycle Station Dynamic Planning System}
\label{section4}
Based on bicycle drop-off location clustering, we established a series of bicycle station graph models and further created a graph sequence model.
In this section, we propose a Bicycle Station Dynamic Planning (BSDP) system to predict the location of stations and their bicycle-sharing demands in the next period.
The GGNN model is used to train the bicycle station graph sequence model and predict a bicycle station graph in the next period.
In addition, we fine-tune the location of bicycle stations by matching the predictions with the government's urban management plan.
Finally, we provide a bicycle station layout recommendation for the DL-BPS system.

\subsection{GGNN-Based Bicycle-station Location Prediction}
\begin{figure}[!ht]
\centering
\includegraphics[width=5.0in]{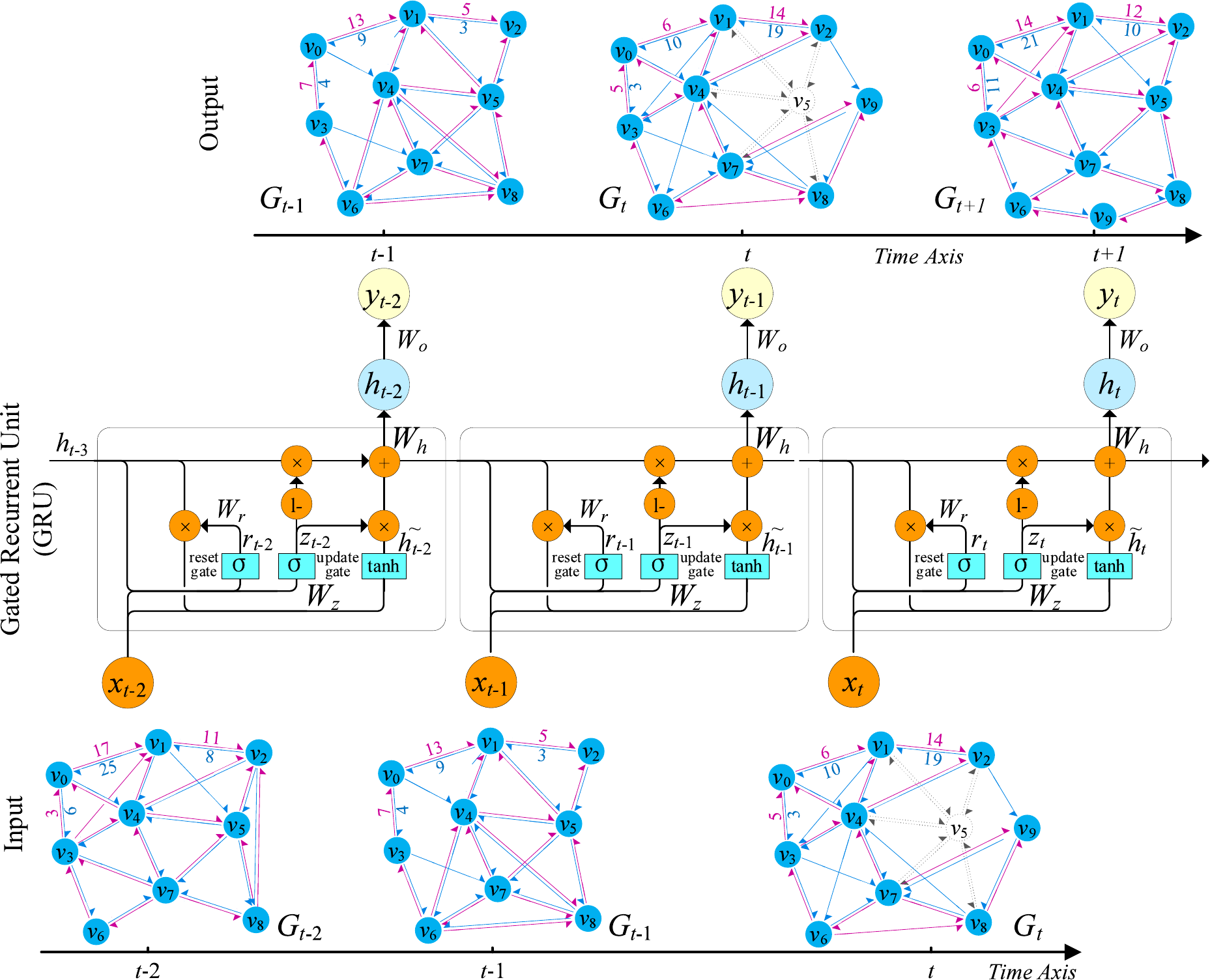}
\caption{Structure of the GGNN-based bicycle-station location prediction model.}
\label{fig07}
\end{figure}

We introduce the Gated Graph Neural Network (GGNN) model \cite{gnn08} to train the large-scale historical graph sequence datasets of each city and predict the bicycle station layout in the next time period.
The structure of the GGNN-based bicycle-station location prediction model is shown in Fig. \ref{fig07}.

\begin{itemize}
\item Input:
    We collect large-scale historical cycling trajectory records from the DL-PBS network in different cities and administrative regions, and then construct a set of bicycle station graph datasets $GS = \{G_{1}, \dots, G_{t}\}$ for each of them.
     Each bicycle station graph $G_{t} \in GS$ in a historical time period is used as an input of the GGNN model.
\item Output:
     Given an input $G_{t}$, the output of the GGNN model is a predicted bicycle station graph $G_{t+1}$ for the $(t+1)$-th time period.
    From $G_{t+1}$, we can obtain the location of the predicted bicycle stations and the number of bicycles needed at each station.
\end{itemize}

(1) The forward-propagation prediction process.

We represent the bicycle station graph sequence dataset $GS = \{G_{1}, \dots, G_{t}\}$ as $X = \{x_{1}, \dots, x_{t}\}$, and use each graph $x_{t} = G_{t}$ as the input of the GGNN model.
We use the Gated Recurent Unit (GRU) cells as the gate layer of the GGNN model.
Let $H=\{h_{1}, \dots, h_{t}\} \in \mathbb{R}^{D \times n}$ be the hidden state matrix of the GRU module, where $D$ is the dimension of the hidden state of each unit.
For the input $x_{t}$ of the $t$-th time period, the values of the reset gate and update gate are calculated as:
\begin{equation}
\label{eq08}
%\begin{aligned}
r_{t} = \sigma\left(W_{r} \odot [h_{t-1}, x_{t}]+ b_{r}\right),
\end{equation}
\begin{equation}
z_{t} = \sigma\left(W_{z} \odot [h_{t-1}, x_{t}] + b_{z}\right),
%\end{aligned}
\end{equation}
where $W_{r}$ and $b_{r}$ are the weight-parameter and bias matrices of the reset gate, $W_{z}$ and $b_{z}$ are those of the update gate, and operation $\odot$ indicates an element-wise multiplication.
$\sigma()$ is a sigmoid activation function, and $\sigma(x) = {(1 + e^{-x})}^{-1}$.
Based on the reset and update gates, we further calculate the values of the hidden and output layers:
\begin{equation}
\label{eq09}
%\begin{aligned}
\widetilde{h}_{t} = \text{tanh} \left(W_{\widetilde{h}} \odot [r_{t} \times h_{t-1}, x_{t}]\right),
\end{equation}
\begin{equation}
h_{t} = (1-z_{t}) \times h_{t-1} + z_{t} \times \widetilde{h}_{t}, 
\end{equation}
\begin{equation}
y_{t} = \sigma\left(W_{o} \odot h_{t}\right).
%\end{aligned}
\end{equation}
In this way, for each input $x_{t} = G_{t}$, we can obtain the corresponding output $y_{t}$ from the GGNN model and treat it as the predicted bicycle station graph $G_{t+1}$ for the $(t+1)$-th time period.

(2) The backward-propagation training process.

In the previous process, we obtain $y_{t}=G_{t+1}$ for each input $x_{t}=G_{t}$ in each period $t$.
Benefiting from large-scale historical bicycle station graph sequence models $G_{1} \sim G_{t}$, we can continuously train the GGNN model to be stable and convergent by comparing the predicted graph value of each historical period with the actual value.
Let $y_{t}$ be the predicted graph and $y_{d}$ be the actual graph of the bicycle stations in the $(t+1)$-th time period.
The core loss functions between different gates and layers are calculated as follows:
\begin{equation}
\label{eq10}
%\begin{aligned}
\delta_{y,t} = (y_{d} - y_{t}) \odot \sigma',
\end{equation}
\begin{equation}
\delta_{h,t} = \delta_{y,t}W_{o} +\delta_{z,t+1}W_{zh}+\delta_{t+1}W_{\widetilde{h}h}r_{t+1} +
\delta_{h,t+1}W_{rh}+\delta_{h,t+1}(1-z_{t+1}),
\end{equation}
\begin{equation}
\delta_{z,t} = \delta_{t,h}(\widetilde{h} - h_{t-1}) \odot \sigma',
\end{equation}
\begin{equation}
\delta_{t} = \delta_{h,t} \odot z_{t} \odot \tanh',
\end{equation}
\begin{equation}
\delta_{r,t} = \left[(\delta_{h,t} \odot z_{t} \odot \tanh')W_{\widetilde{h}h}\right] \odot \sigma',
%\end{aligned}
\end{equation}
where $W_{rx}$ is the weight matrix between the reset gate and the input layer, $W_{rh}$ is the weight matrix between the reset gate and the previous hidden layer, and $W_{r}$ is the join link of $W_{rx}$ and $W_{rh}$:
\begin{equation}
\label{eq11}
%\begin{aligned}
W_{r} = W_{rx} + W_{rh},
\end{equation}
\begin{equation}
W_{z} = W_{zx} + W_{zh},
\end{equation}
\begin{equation}
W_{\widetilde{h}} = W_{\widetilde{h}x} + W_{\widetilde{h}h}.
%\end{aligned}
\end{equation}
We use the above equations to iteratively update the weight matrixes, such as $W_{r}$, $W_{z}$, $W_{\widetilde{h}}$, $W_{h}$, and $W_{o}$.
In this way, the weight parameters between different gates and layers are successively updated to obtain a stable and
convergent GGNN model.
The detailed steps of GGNN-based bicycle station prediction are presented in Algorithm \ref{alg3}.

\begin{algorithm}[!ht]
\caption{GGNN-based bicycle station prediction}
\label{alg3}
\begin{algorithmic}[1]
\REQUIRE ~\\
    $GS = \{G_{1}, \dots, G_{t}\}$:  A historical graph sequence model of DL-BPS bicycle stations;\\
\ENSURE ~\\
    $G_{t+1}$: the predicted bicycle station graph for the $(t+1)$-th time period.
\STATE initialize the weight parameters of the GGNN model;
\FOR {each iteration epoch $i$}
\FOR {each graph dataset $G_{t}$ in $GS$}
\STATE  forward propagation and predict $y_{t} \leftarrow \text{GGNN}(G_{t})$;
\STATE  calculate the loss functions using Eq. (\ref{eq10});
\STATE  update the weight parameters of the GGNN model using Eq. (\ref{eq11});
\ENDFOR
\ENDFOR
\STATE save the trained GGNN model;
\STATE predict bicycle station graph $G_{t+1} \leftarrow$ GGNN($GS$);
\RETURN $G_{t+1}$.
\end{algorithmic}
\end{algorithm}

\subsection{Bicycle Station Layout Recommendation}
Based on the prediction results of bicycle-sharing demand, we can obtain the location of bicycle stations and the number of bicycles needed at each station in the next period.
As mentioned above, in a DL-PBS network, users can not only find available bicycles at any nearest location via GPS positioning, but also return bicycle to anywhere near their destination.
Given that users may return public bicycles to locations that are not permitted by the government, the location of bicycle stations predicted from these records may also serve as illegal.
Therefore, we need to fine-tune the location of bicycle stations by matching the predictions with the government's urban management plan.
Specifically, if a predicted station is in a permitted area, set it as a bicycle station, otherwise, we fine-tune the location of the current station to the nearest permitted area.

Let $G_{t+1} = (V, E, D, W)$ be a predicted bicycle station graph model for the subsequent period $t+1$, where $V=\{v_{1}, v_{2}, \dots, v_{n}\}$ represents a set of candidate bicycle stations.
Each station $v_{i}=(\varphi_{i}, \psi_{i}, n_{i})$ consists of three attributes: the station location (longitude $\varphi_{i}$ and latitude $\psi_{i}$) and the number $n_{i}$ of bicycles needed at this station.
Let $P=\{p_{1}, p_{2}, \dots, p_{m}\}$ be a set of urban public places that allow parking of public bicycles (termed as legal parking positions).
Each position $p_{j}=(\varphi_{j}, \psi_{j}, n_{j})$ also contains three attributes: the location (longitude $\varphi_{j}$ and latitude $\psi_{j}$) and the maximum number $n_{j}$ of bicycles that the position can hold.
For each predicted bicycle station $v_{i}$, we find the nearest legal parking position $p_{j} \in P$.
There are three situations between $v_{i}$ and $p_{j}$.
\begin{itemize}
\item Case (a): $d_{ij} \leq \theta_{d}$ and $n_{j} \geq n_{i}$: $v_{i}$ is in a legal parking position with sufficient space.
    We use Eq. (\ref{eq03}) to obtain the distance $d_{ij}$ between $v_{i}$ and $p_{j}$.
    Considering the deviation of GIS data acquisition, we set a deviation threshold $\theta_{d}$ for position matching.
    If $d_{ij} \leq \theta_{d}$, the locations of $v_{i}$ and $p_{j}$ are considered coincident, that is, $v_{i}$ is in $p_{j}$.
    In addition, if $n_{j} \geq n_{i}$, this means that $p_{j}$ has sufficient space to accommodate $v_{i}$'s  predicted bicycles, and we accept this station and the predicted bicycles without any adjustments.
\item Case (b): $d_{ij} \leq \theta_{d}$ and $n_{j} < n_{i}$: $v_{i}$ is in a legal parking position with insufficient space.
    If $d_{ij} \leq \theta_{d}$ and $n_{j} < n_{i}$, it means that the number of bicycles at $v_{i}$ exceeds the available space provided by $p_{j}$.
    We need to split the station $v_{i}$ by finding another near legal parking position $p_{j}'$ for the extra bicycles ($n_{i} - n_{j}$).
\item Case (c): $d_{ij} > \theta_{d}$: $v_{i}$ is in an illegal position.
    If $v_{i}$ is not in any legal parking position, we need to fine-tune the location of $v_{i}$ to $p_{j}$: $(\varphi_{i}, \psi_{i}) = (\varphi_{j}, \psi_{j})$.
    Then we go to Case (a) or Case (b) to further judge whether the available space of $p_{j}$ meets the conditions.
\end{itemize}

\begin{figure}[!ht]
\centering
\includegraphics[width=3.5in]{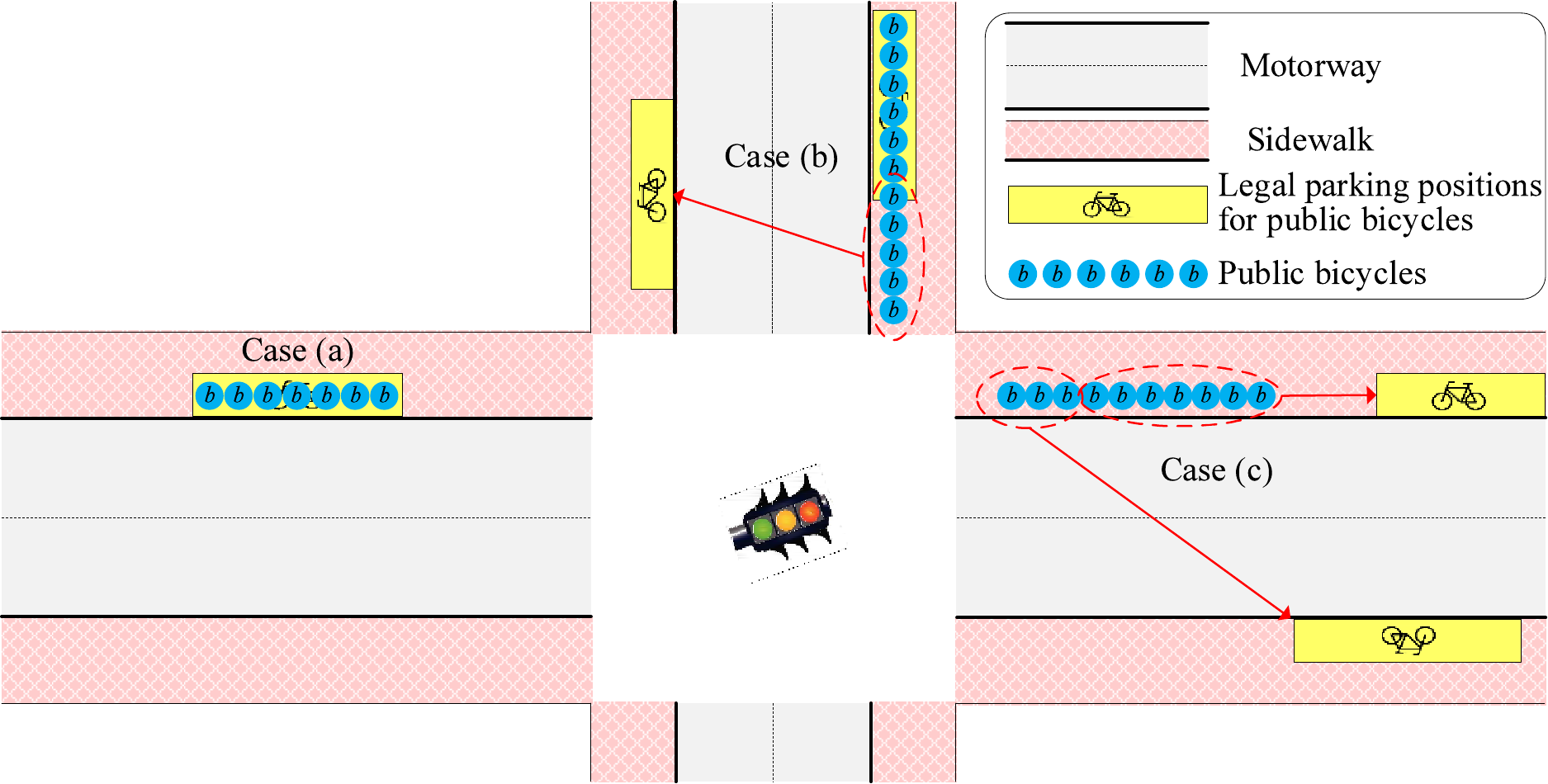}
\caption{Examples of the relationship between predicted bicycle stations and legal parking positions.}
\label{fig08}
\end{figure}

Examples of the relationship between predicted bicycle stations and legal parking positions are shown in Fig. \ref{fig08}.

After matching the predicted bicycle stations to the legal parking positions, we fine-tune the stations by updating the predicted station locations and the number of bicycles at each station.
Finally, we obtain the recommendation scheme of bicycle station layout for the current urban area.
Algorithm \ref{alg4} gives detailed steps of the fine-tuning and recommendation process of the predicted bicycle stations.

\begin{algorithm}[!ht]
\caption{Bicycle station fine-tuning and recommendation}
\label{alg4}
\begin{algorithmic}[1]
\REQUIRE ~\\
    $G_{t+1}$: The predicted graph model of bicycle stations;\\
    $P$: A set of legal parking positions for public bicycles;\\
    $\theta_{d}$ : Distance deviation threshold of station location matching;\\
\ENSURE ~\\
    $G_{t+1}'$: The fine-tuned bicycle station layout scheme.
\STATE load predicted bicycle stations $V$ from $G_{t+1}$;
\WHILE {$V$ is not empty}
\FOR {each station $v_{i}$ in $V$}
\STATE find the nearest legal parking position $p_{j} \leftarrow (v_{i}, P)$;
\IF {$d_{ij} \leq \theta_{d}$}
\IF {$n_{j} \geq n_{i}$}
\STATE move $v_{i}$ from $G_{t+1}$ to $G_{t+1}'$;
\ELSE
\STATE update the value of vertex $v_{i}$ as ($\varphi_{i}, \psi_{i}, n_{j}$);
\STATE move $v_{i}$ from $G_{t+1}$ to $G_{t+1}'$;
\STATE find another nearest legal parking position  $p_{j}' \leftarrow (v_{i}, P-p_{j})$;
\STATE create a new vertex $v_{i}'$ with the attributes of ($\varphi_{j'}$, $\psi_{j'}$, $n_{i}-n_{j}$);
\STATE append $v_{i}'$ to $G_{t+1}'$;
\ENDIF
\ELSE
\STATE update the location of station $v_{i}$ ($\varphi_{i}$, $\psi_{i}$) $\leftarrow$ ($\varphi_{j}$, $\psi_{j}$);
\ENDIF
\ENDFOR
\ENDWHILE
\RETURN $G_{t+1}'$.
\end{algorithmic}
\end{algorithm}

\section{Experiments}
\label{section5}

\subsection{Experimental Setup}
We collect bicycle GPS datasets and cycling trajectory records from a DL-PBS provider in China.
These datasets are collected from 16 administrative regions in Beijing, China from Jan. 1st, 2018 to Dec. 31st, 2019.
According to administrative divisions, these datasets are divided into 16 spatial subsets, and each spatial subset is further divided into multiple temporal subsets by days or weeks.
The demand for shared bikes is real-time, and bicycle demand forecasting and station planning are the basis of bicycle dispatching. The total circulation and dispatch frequency of bicycles in different administrative regions are different. In this case, we forecast bicycle demand according to the time period of bicycle dispatching (i.e., daily or weekly).

In the comparison experiments, we use k-fold cross validation to divide the training set and test set.
We set $k = 5$ and perform 5-fold cross validation.
For each spatial subset of an administrative division, we divide the spatio-temporal subsets into 5 groups.
Then, 4 groups are use as the training set and the remaining 1 group is used as the test set.
We repeat this process until every 5-fold serve as the test set.
Finally, we take the average of the recorded scores as the accuracy of the corresponding algorithm.

\begin{figure}[!ht]
\centering
\subfigure[Clustering results (October 22, 2018)]{\includegraphics[width=2.7in]{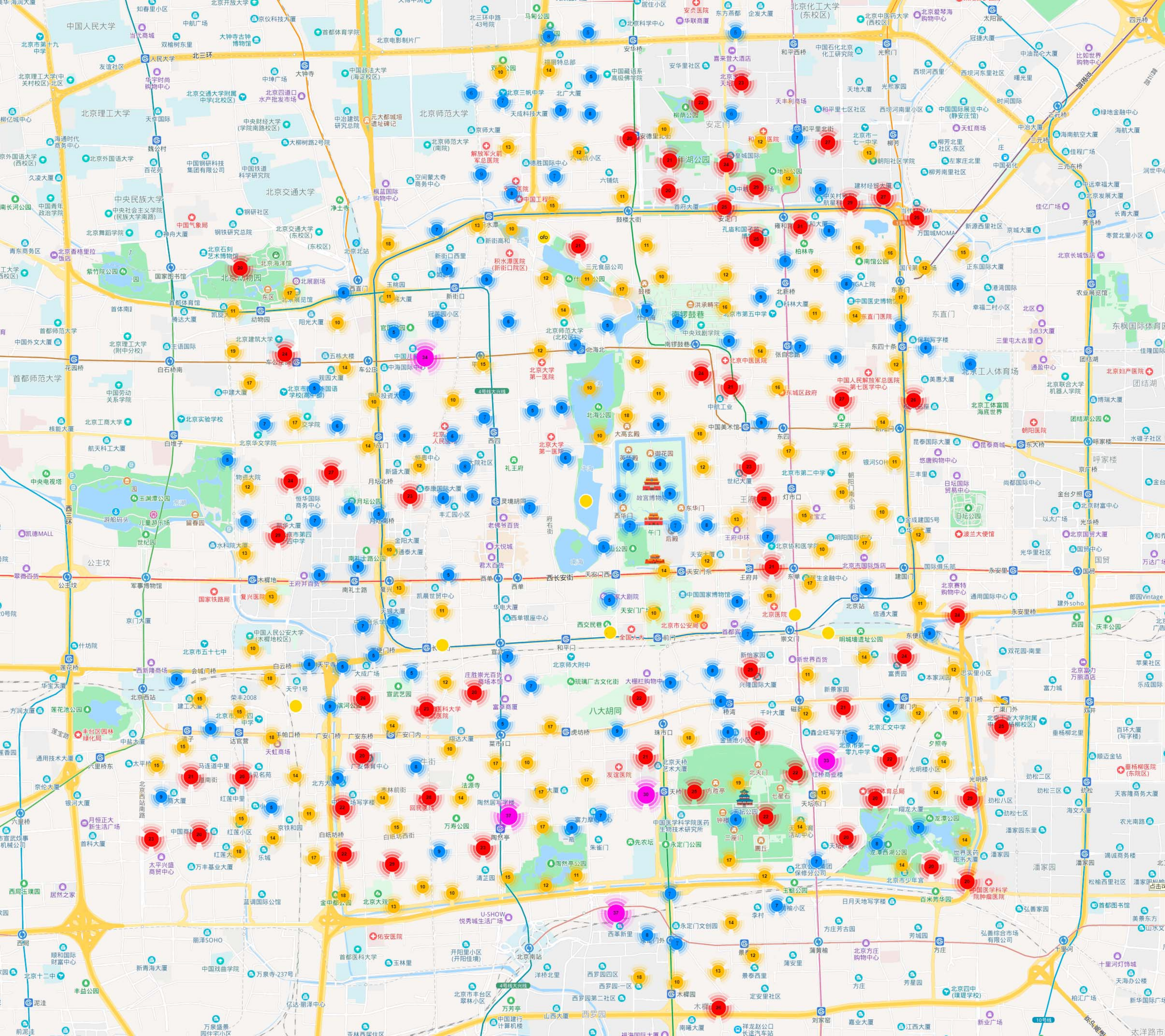}}
 \subfigure[Graph model (October 22, 2018)]{\includegraphics[width=2.7in]{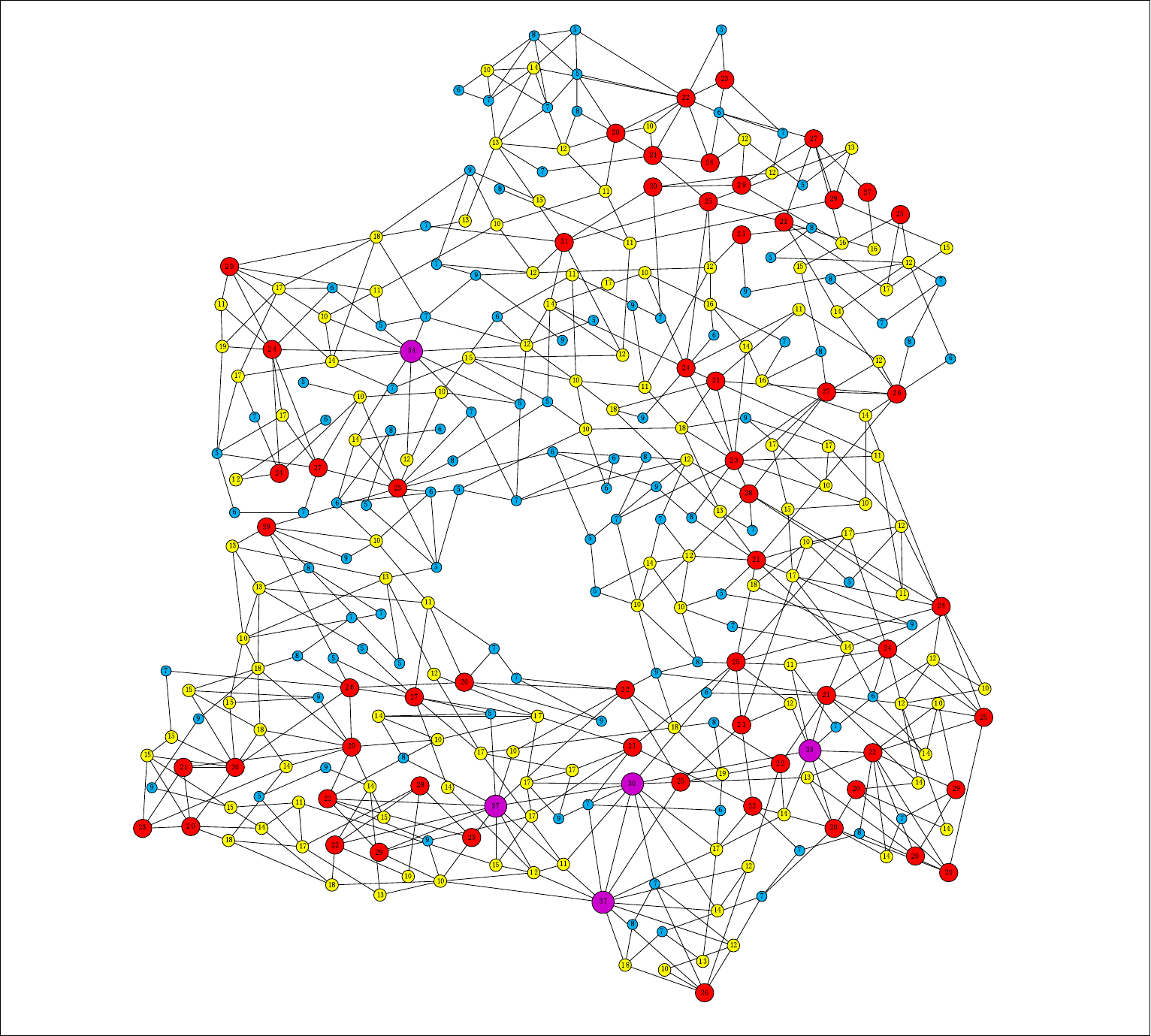}}
 \subfigure[Graph model (October 23, 2018)]{\includegraphics[width=2.7in]{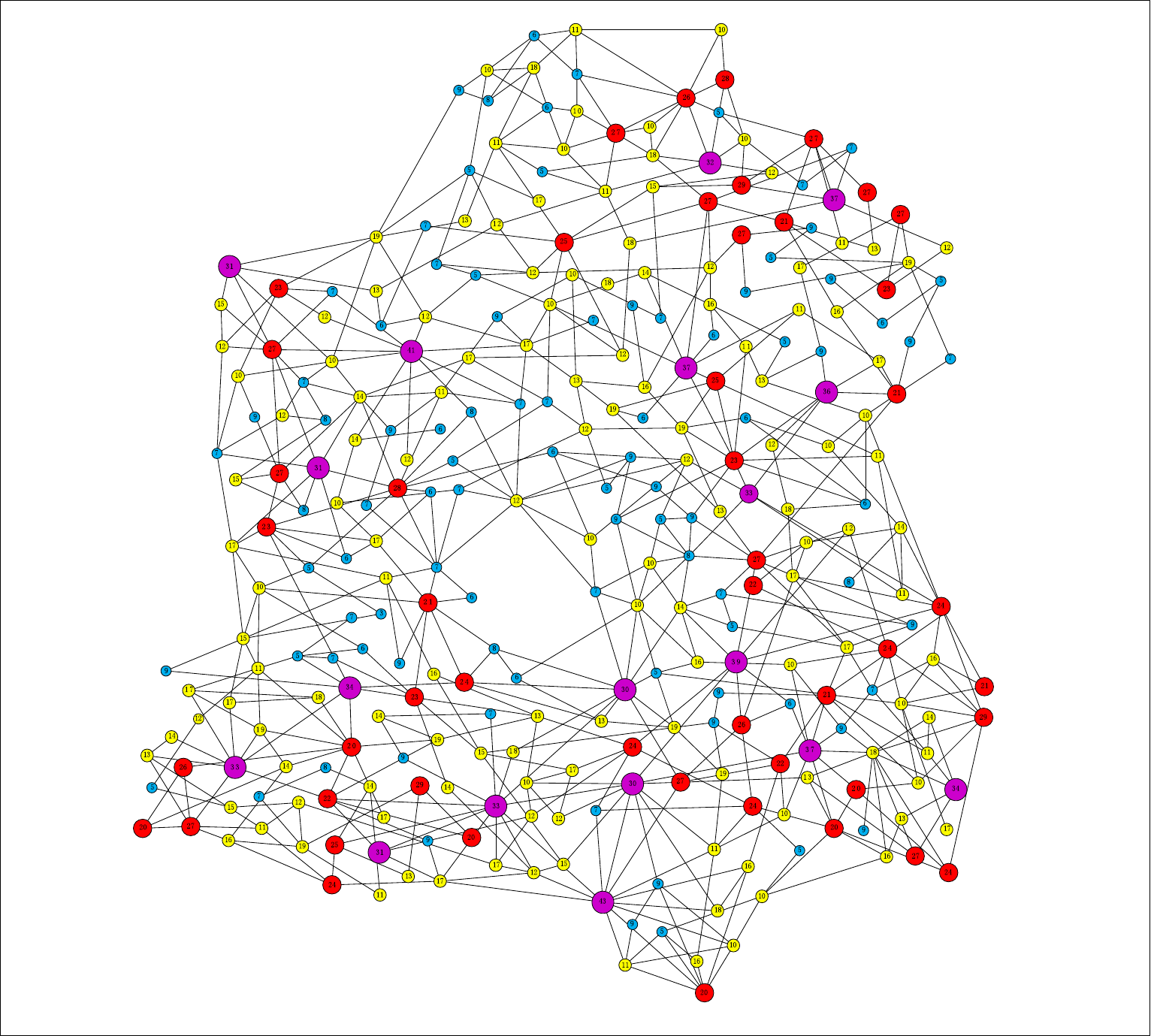}}
 \subfigure[Predicted graph model (October 24, 2018)]{\includegraphics[width=2.7in]{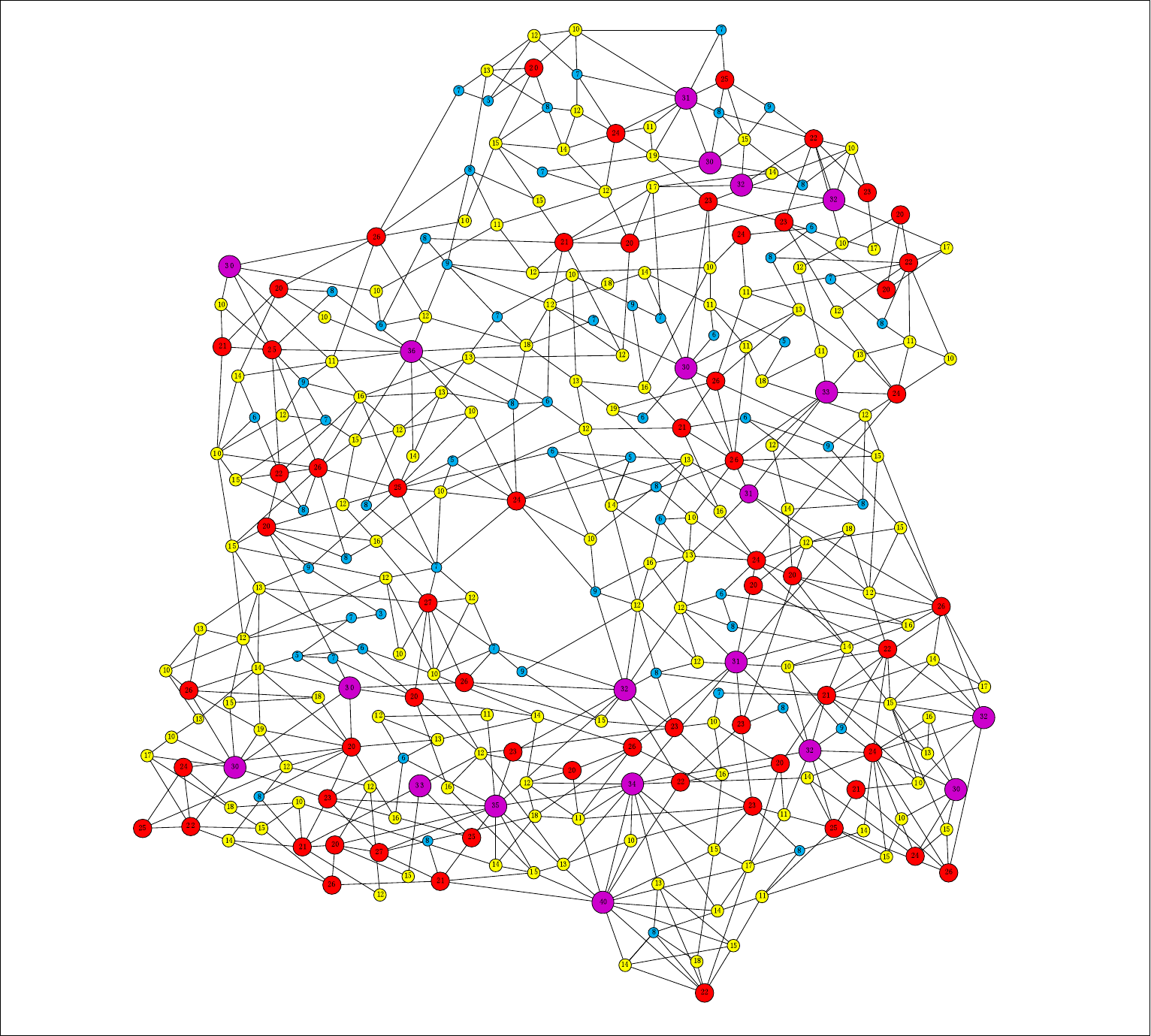}}
 \caption{Results of bicycle drop-off location clustering and bicycle station prediction in the Dongcheng and Xicheng Districts.
 (a) is the clustering results of bicycle drop-off locations on October 22, 2018, where four types of bicycle stations are detected, such as micro stations, small stations, medium stations, and large stations.
 (b) is the bicycle station graph model based on the clustering results on October 22, 2018.
 (c) is the bicycle station graph model based on the clustering results on October 23, 2018.
 (d) is the predicted bicycle-station graph for October 24, 2018.}
 \label{fig09}
\end{figure}

\subsection{Discussion of Experimental Results}
In this section, we discuss the experiment results of bicycle drop-off location clustering and bicycle-station location prediction.
In three cases, we discuss the experimental results of four administrative regions in Beijing, such as Dongcheng District, Xicheng District, Haidian District, and Fengtai District.

\subsubsection{Case Study 1: Dongcheng and Xicheng Districts}
In the first case, we divide the spatial DL-PBS datasets of Dongcheng and Xicheng Districts into 730 temporal subsets by day, each of which has approximately 418,392 cycling trajectory records.
We implement the proposed bicycle drop-off location clustering algorithm on each historical temporal subset and obtain the corresponding clustering results.
Based on the clustering result of each temporal subset, we construct a corresponding bicycle-station graph model for each time period, and then collect all graph models in different periods to form a graph sequence model.
Finally, we execute the GGNN algorithm on the graph sequence model to predict the bicycle-station graph model in the next time period.
Taking the graph models from October 22, 2018 to October 24, 2018 as an example, the experimental results are shown in Fig. \ref{fig09}.

Fig. \ref{fig09}(a) shows the clustering results of bicycle drop-off locations in Dongcheng and Xicheng Districts during October 22, 2018.
According to the number of bicycles that can be accommodated at each bicycle station, we divide these stations into four levels: micro station (each station can accommodate 5 to 10 bicycles), small station (10 to 20 bicycles), medium station (20 to 30 bicycles), and large station (more than 30 bicycles).
In this way, from the clustering results, we obtain 145 micro stations, 137 small stations, 59 medium stations, and 5 large stations.
Bicycle drop-off location clusters with less than 5 bicycles will be ignored.

Based on the clustering results, we include the cycling records between the clusters and build a bicycle-station graph model for October 22, 2018.
Then we calculate the revenue and utility of each candidate station and remove 43 inferior stations from the graph $G_{20181022}$, as shown in Fig. \ref{fig09}(b).
In the same way, we build a bicycle-station graph model for October 23, 2018 based on the corresponding clustering results.
As shown in Fig. \ref{fig09}(c), there are 91 micro stations, 136 small stations, 49 medium stations, and 18 large stations in the graph model $G_{20181023}$.
By comparing the graphs $G_{20181022}$ and $G_{20181023}$, we can see that the number of bicycles at almost every station has changed dynamically.
From $G_{20181022}$ and $G_{20181023}$, 6 micro stations and 2 small stations disappeared, but new 2 micro stations appeared.
As bicycles flow, 13 micro stations expand to small stations.
6 small stations expand to medium stations and 4 shrink to micro stations.
13 medium stations expand to large stations and 3 shrink to small stations.
In this way, we continue to perform the processes of bicycle drop-off location clustering and bicycle-station graph modeling for each temporal subset, and create a graph sequence model.

After obtaining the graph sequence model of Dongcheng and Xicheng Districts, we train the GGNN model using the graph sequence data and get a predicted bicycle-station graph $G_{20181024}$ in the next day (October 24, 2018), as shown in Fig. \ref{fig09}(d).
Then, we fine-tune the location of bicycle stations by matching the predictions with Dongcheng and Xicheng Districts' urban management plan.
Finally, we get the location of each bicycle station and the number of bicycles needed.

\subsubsection{Case Study 2: Haidian District}
\begin{figure}[!ht]
\centering
\subfigure[Clustering results (The 1st week of June 2018)]{\includegraphics[width=2.7in]{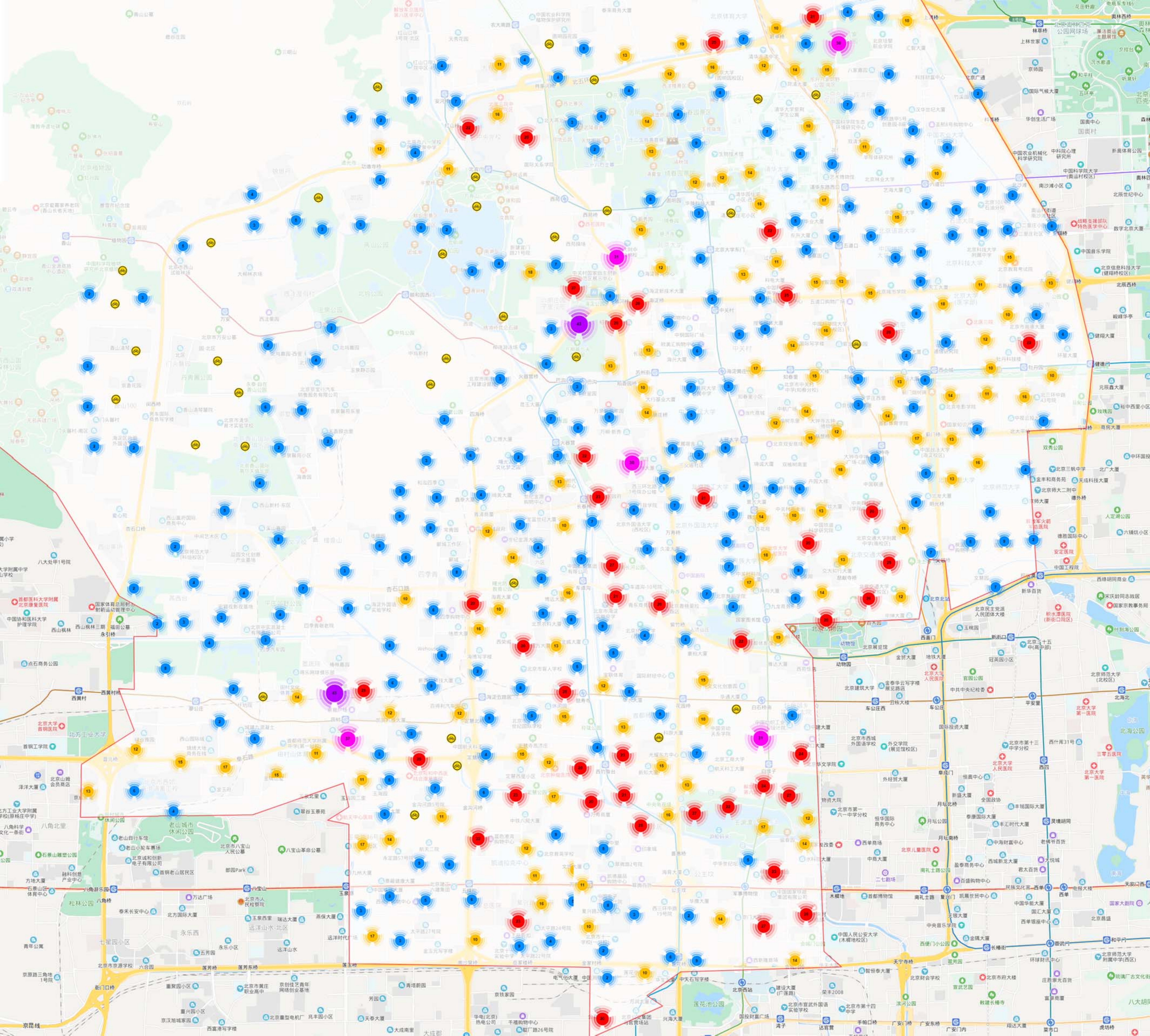}}
 \subfigure[Graph model (The 1st week of June 2018)]{\includegraphics[width=2.7in]{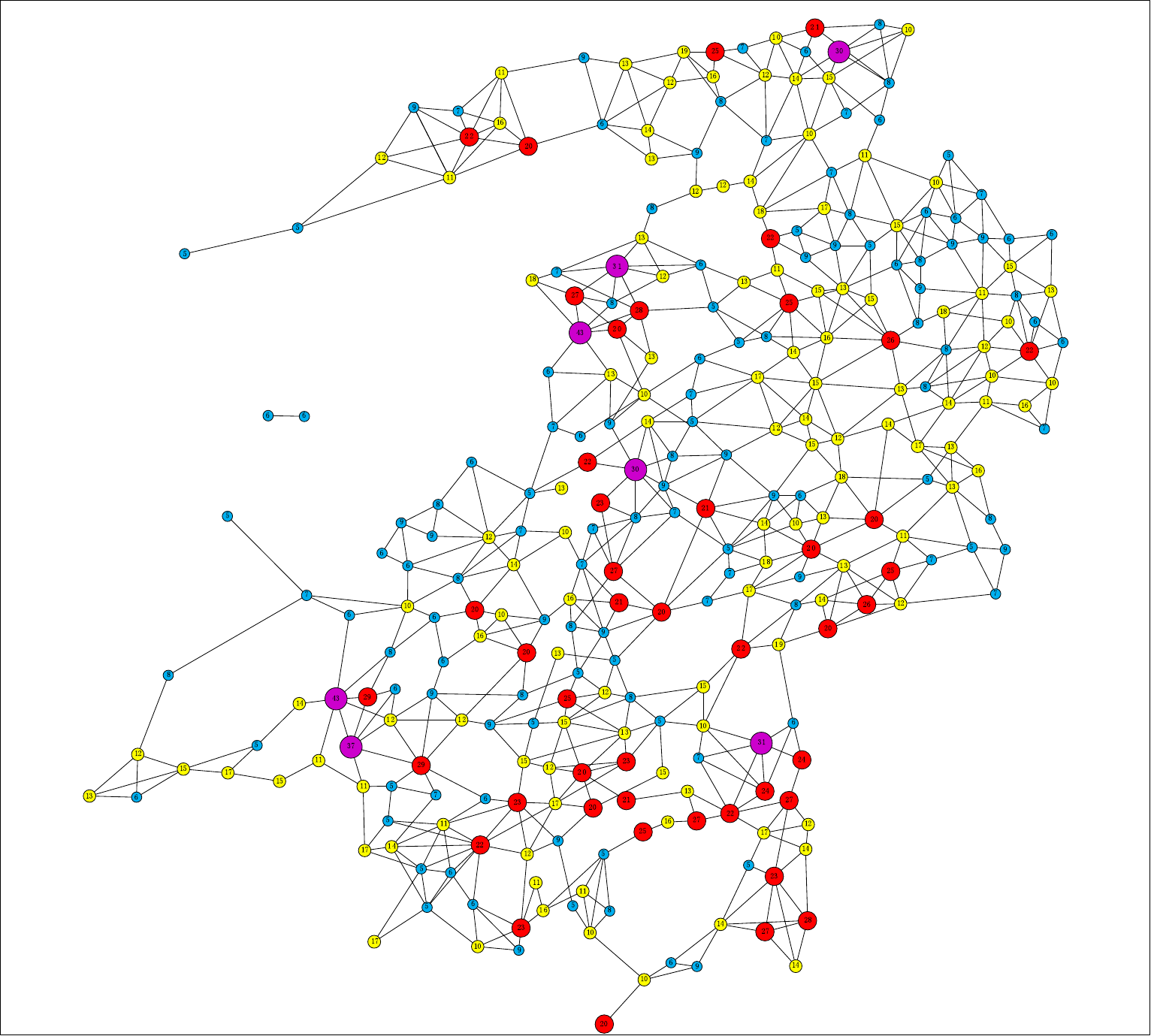}}
 \subfigure[Graph model (The 2nd week of June 2018)]{\includegraphics[width=2.7in]{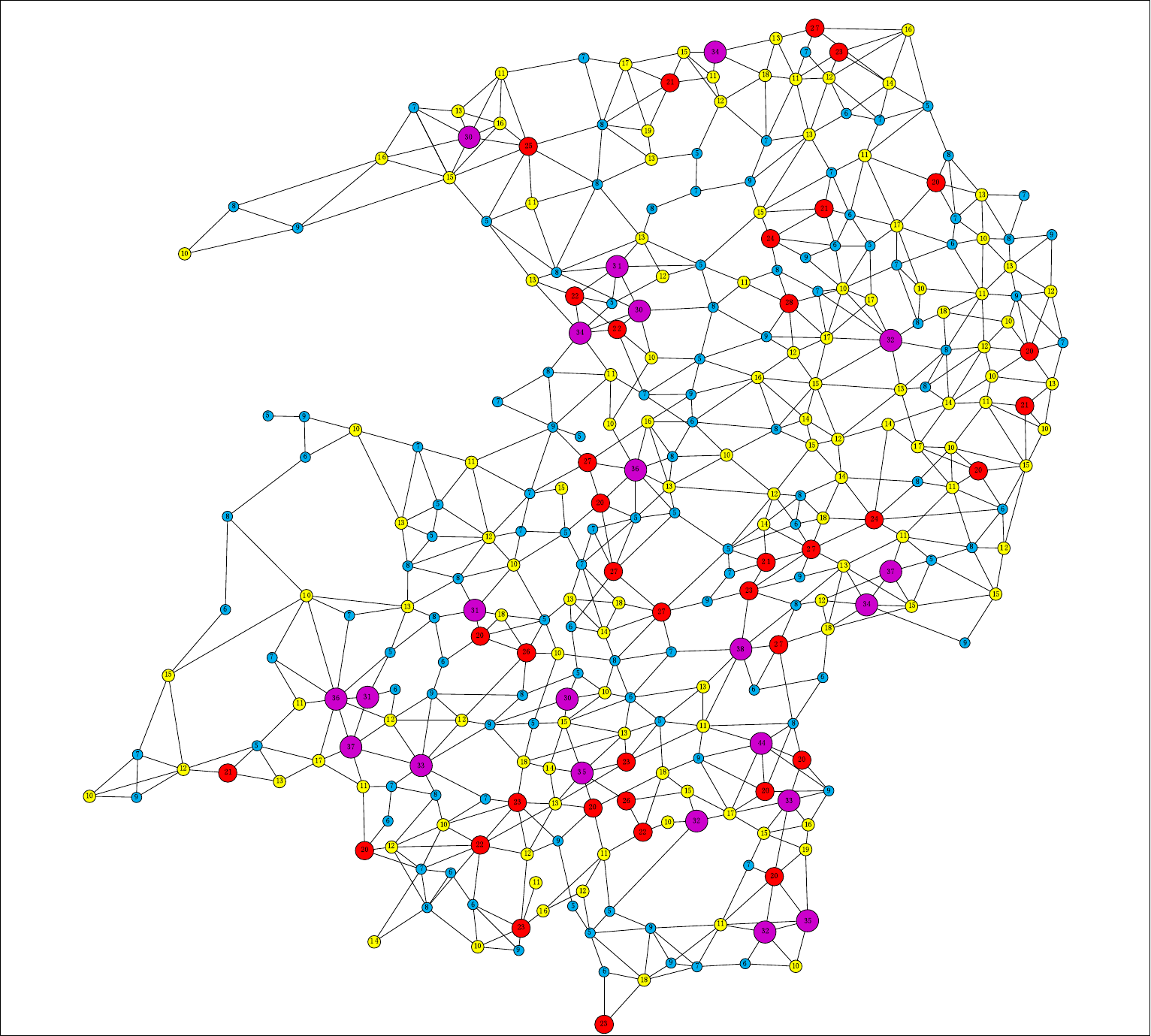}}
 \subfigure[Predicted graph model (The 3rd week of June 2018)]{\includegraphics[width=2.7in]{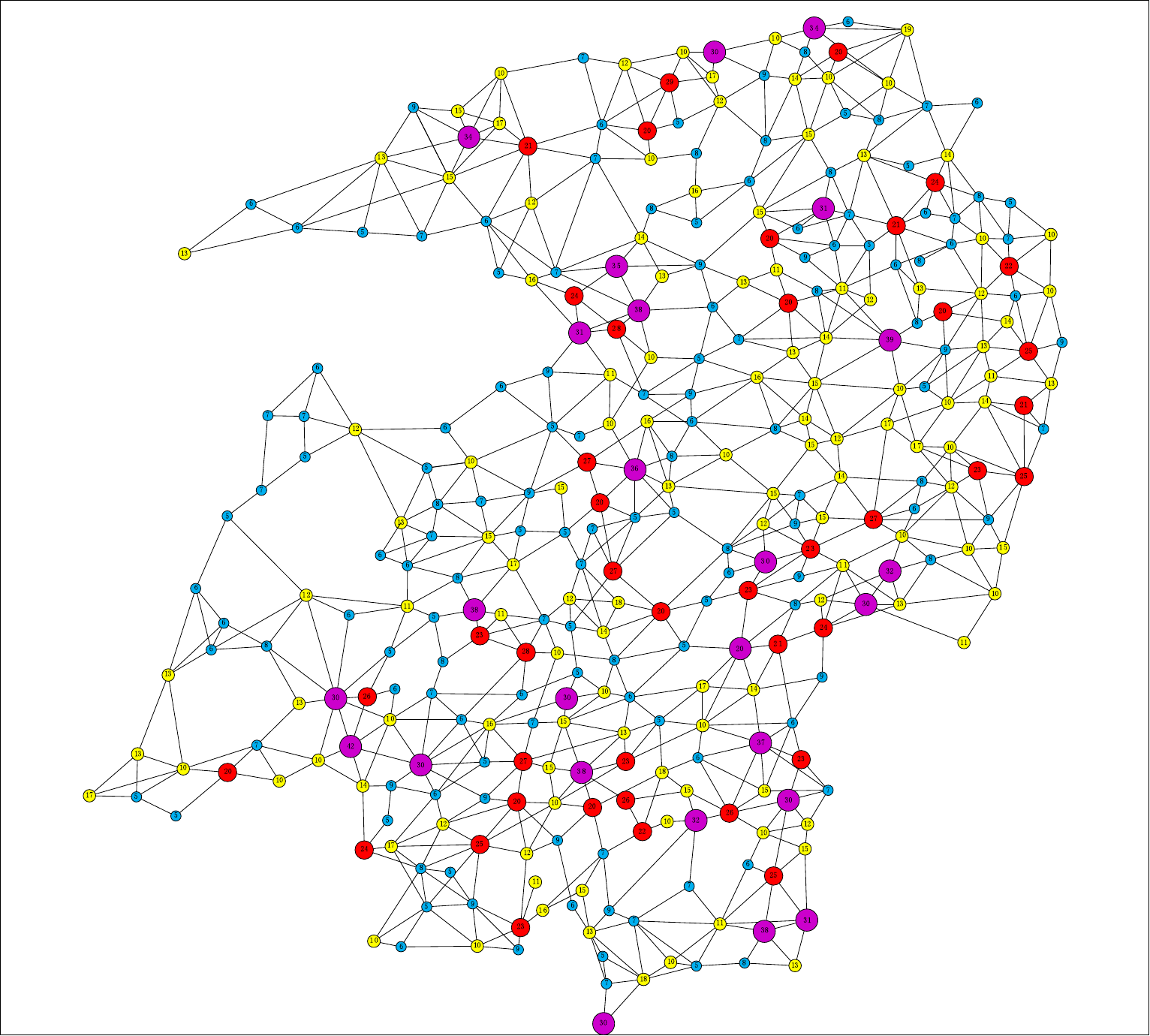}}
 \caption{Results of bicycle drop-off location clustering and bicycle station prediction in the Haidian District. We group the historical bicycle GPS data sets and predict the bicycle station graph by weeks.
 (a) and (b) are the clustering results of bicycle drop-off locations and the related bicycle station graph model for the 1st week of June 2018,
 (c) is the graph model for the 2nd week of June 2018,
 and (d) is the predicted bicycle-station graph for the 3rd week of June 2018.}
 \label{fig10}
\end{figure}

In the second case, we discuss the experimental results of the spatial DL-PBS datasets in Haidian District, Beijing.
The spatial DL-PBS dataset is divided into 104 temporal subsets by week, each of which has approximately 3,521,486 cycling trajectory records.
The experimental results of bicycle drop-off location clustering, bicycle-station graph construction, and bicycle station prediction are shown in Fig. \ref{fig10}.

Fig. \ref{fig10}(a) shows the clustering results of bicycle drop-off locations in Haidian District during the first week of June 2018.
We obtain 123 micro stations, 119 small stations, 45 medium stations, and 7 large stations.
Based on the clustering results, we include the cycling records between the clusters and build a bicycle-station graph model for this period.
Then we calculate the revenue and utility of each candidate station and remove 28 inferior stations from the graph $G_{2018061w}$, as shown in Fig. \ref{fig10}(b).
As shown in Fig. \ref{fig10}(c), we build the bicycle-station graph model $G_{2018062w}$ for the second week of June 2018, which contains 139 micro stations, 121 small stations, 41 medium stations, and 25 large stations.
We continue to perform the processes of bicycle drop-off location clustering and bicycle-station graph modeling for each temporal subset in the same way, and create a graph sequence model.
Next step, we train the GGNN model using the graph sequence data and get a predicted bicycle-station graph $G_{2018063w}$ in the next period (the third week of June 2018), as shown in Fig. \ref{fig10}(d).
After fine-tuning the location of bicycle stations with Haidian District's urban management plan, we get the location of each bicycle station and the number of bicycles needed.

\subsubsection{Case Study 3: Fengtai District}
\begin{figure}[!ht]
\centering
\subfigure[Clustering results (The 1st week of Sept. 2018)]{\includegraphics[width=2.7in]{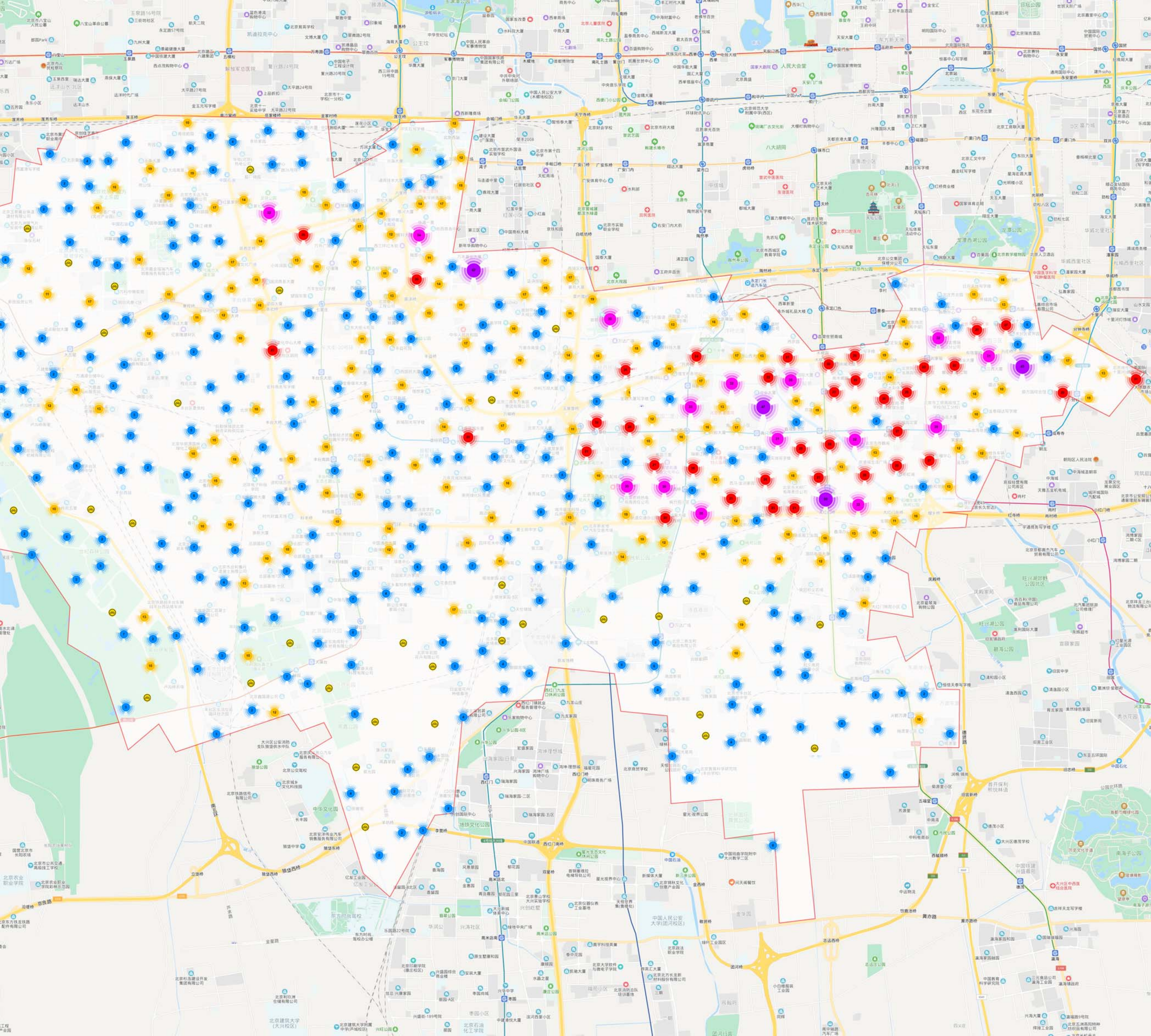}}
 \subfigure[Graph model (The 1st week of Sept. 2018)]{\includegraphics[width=2.7in]{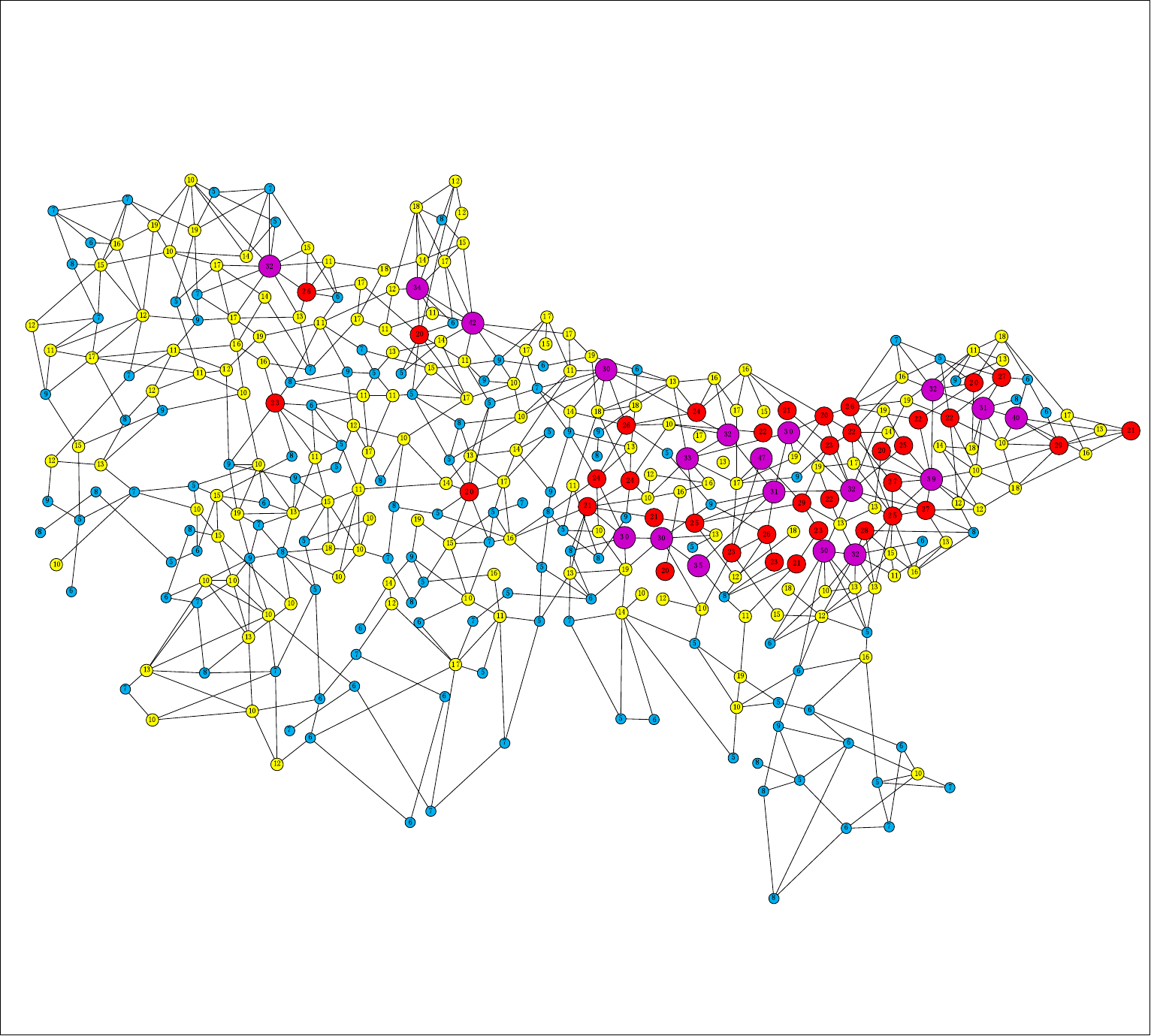}}
 \subfigure[Graph model (The 2nd week of Sept. 2018)]{\includegraphics[width=2.7in]{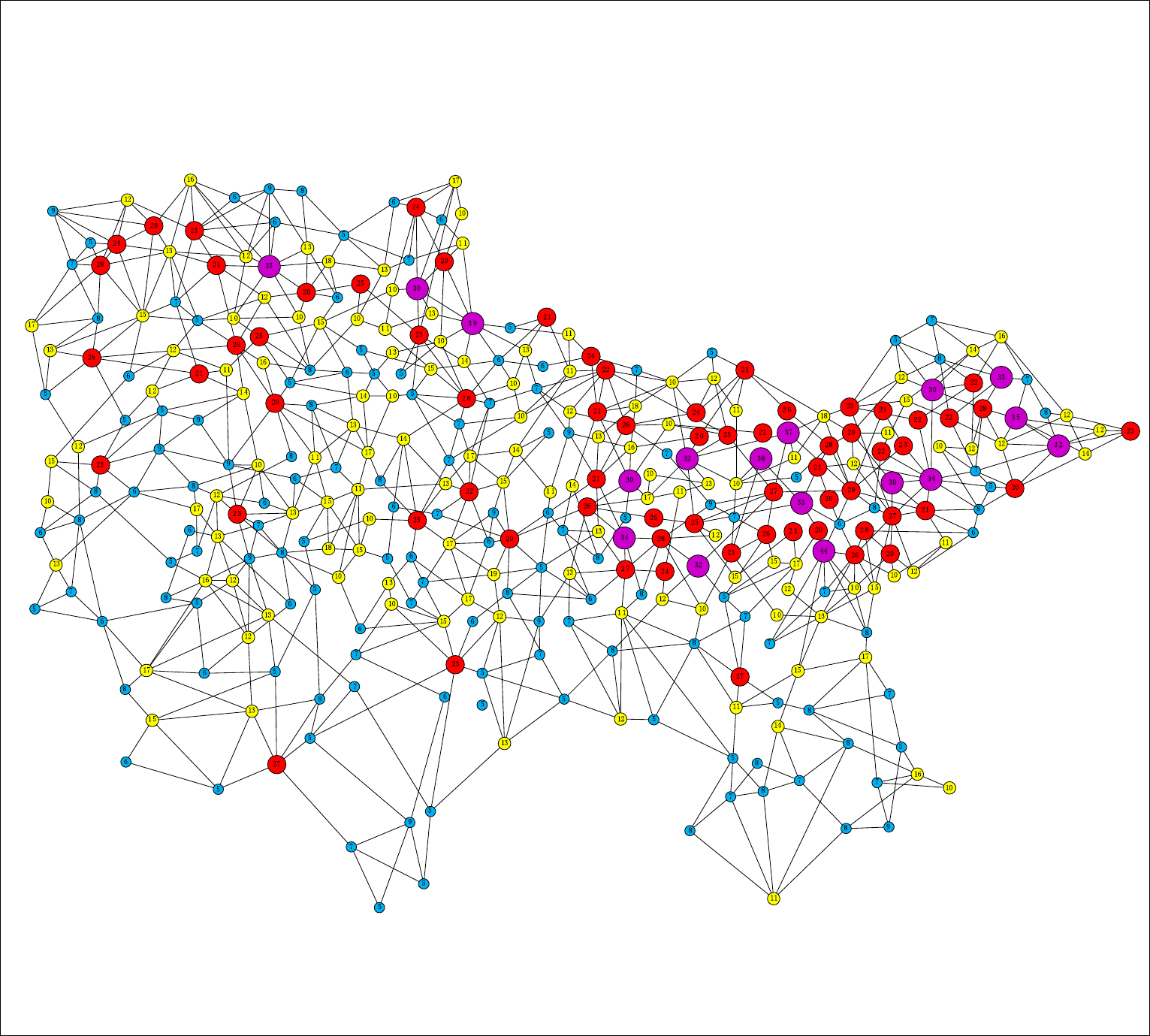}}
 \subfigure[Predicted graph model (The 3rd week of Sept. 2018)]{\includegraphics[width=2.7in]{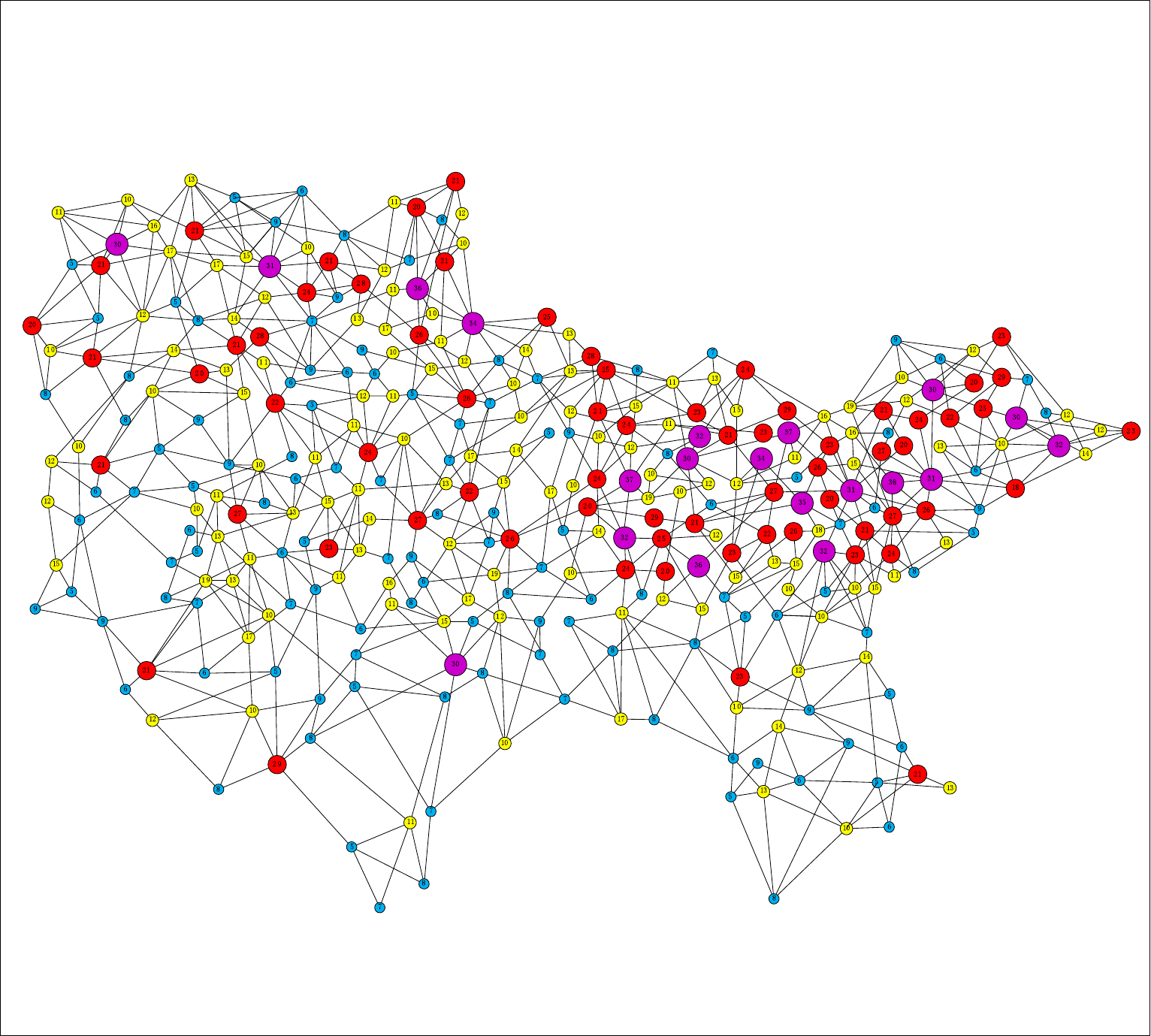}}
 \caption{Results of bicycle drop-off location clustering and bicycle station prediction in the Fengtai District.
 (a) and (b) are the clustering results of bicycle drop-off locations and the related bicycle station graph model for the 1st week of June 2018, where 131 micro stations, 160 small stations, 37 medium stations, and 19 large stations are detected.
 (c) is the graph model for the 2nd week of June 2018, where 141 micro stations, 129 small stations, 67 medium stations, and 17 large stations are detected.
 (d) is the predicted bicycle-station graph for the 3rd week of June 2018, where 119 micro stations, 125 small stations, 67 medium stations, and 20 large stations are predicted.}
 \label{fig11}
\end{figure}

In the third case, we discuss the experimental results of the spatial DL-PBS datasets in Fengtai District, Beijing.
We divide the spatial DL-PBS dataset into 104 temporal subsets by week, each of which has approximately 3,107,753 cycling trajectory records.
The experimental results of bicycle drop-off location clustering, bicycle-station graph construction, and bicycle station prediction are shown in Fig. \ref{fig11}.

Fig. \ref{fig11}(a) shows the clustering results of bicycle drop-off locations in Fentai District during the first week of Sept. 2018.
We obtain 131 micro stations, 160 small stations, 37 medium stations, and 19 large stations.
Based on the clustering results, we include the cycling records between the clusters and build a bicycle-station graph model $G_{2018091w}$ for the current period, as shown in Fig. \ref{fig11}(b).

We continue to build the bicycle-station graph model $G_{2018092w}$ for the second week of Sept. 2018, which contains 141 micro stations, 129 small stations, 67 medium stations, and 17 large stations, as shown in Fig. \ref{fig11}(c).
We continue to perform the above processes in the same way and create a graph sequence model.
Then we train the GGNN model using the graph sequence data and get a predicted bicycle-station graph $G_{2018093w}$ in the next period (the third week of Sept. 2018), as shown in Fig. \ref{fig11}(d).
Note that $G_{2018093w}$ is not only predicted by $G_{2018093w}$ and $G_{2018093w}$, but is predicted by all historical graph models in the current spatial dataset.
After fine-tuning the location of bicycle stations with Fengtai District's urban management plan, we get the location of each bicycle station and the number of bicycles needed.
The experimental results show that by learning a large number of historical graph sequence data, the GGNN model can capture changes in bicycle stations in different periods and accurately predict the bicycle station layout in the next period.

\subsection{Evaluation of Bicycle Drop-off Location Clustering}
\begin{figure}[!ht]
\centering
 \subfigure[Dataset in Dongcheng District]{\includegraphics[width=2.7in]{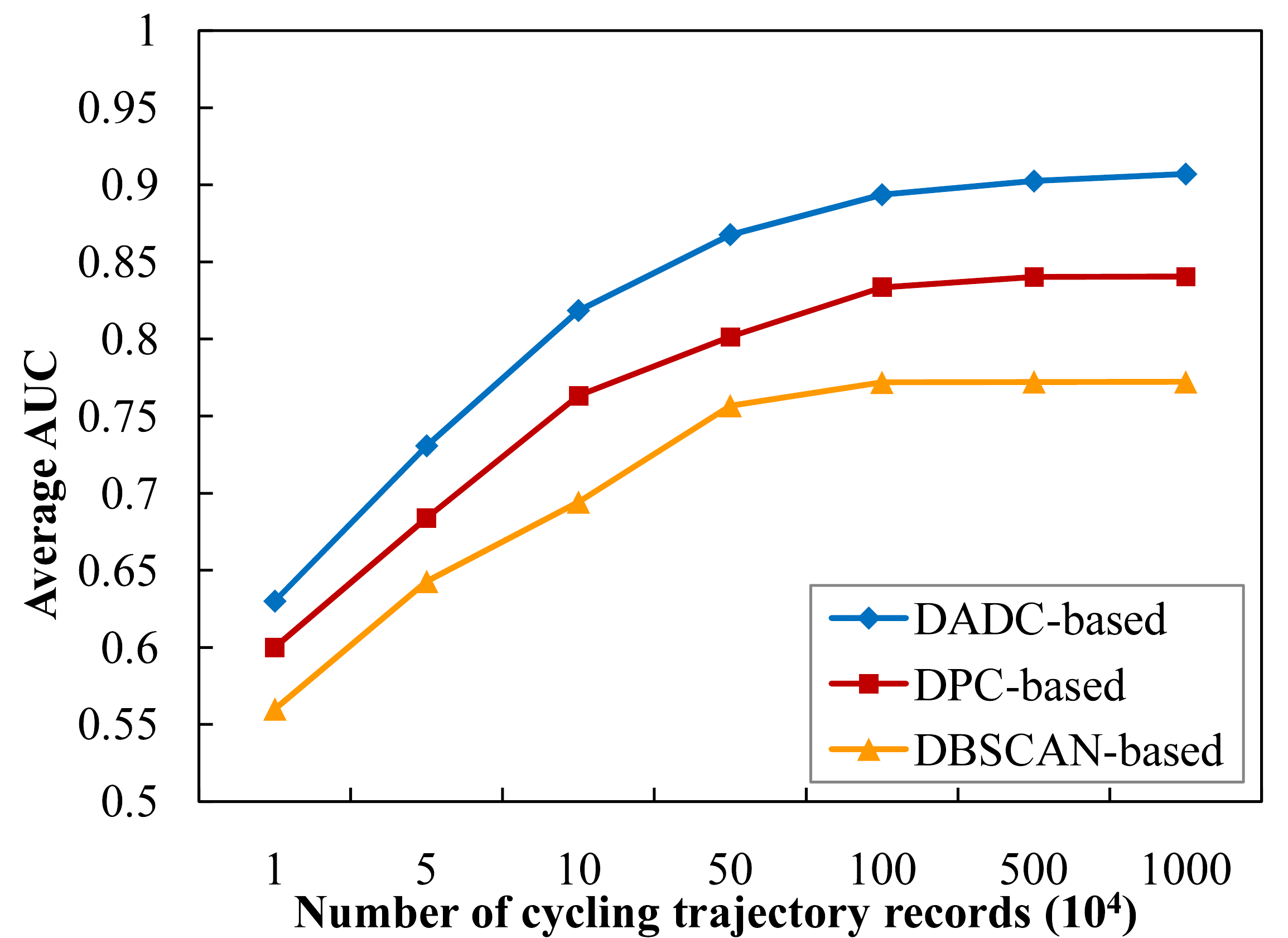}}
 \subfigure[Dataset in Xicheng District]{\includegraphics[width=2.7in]{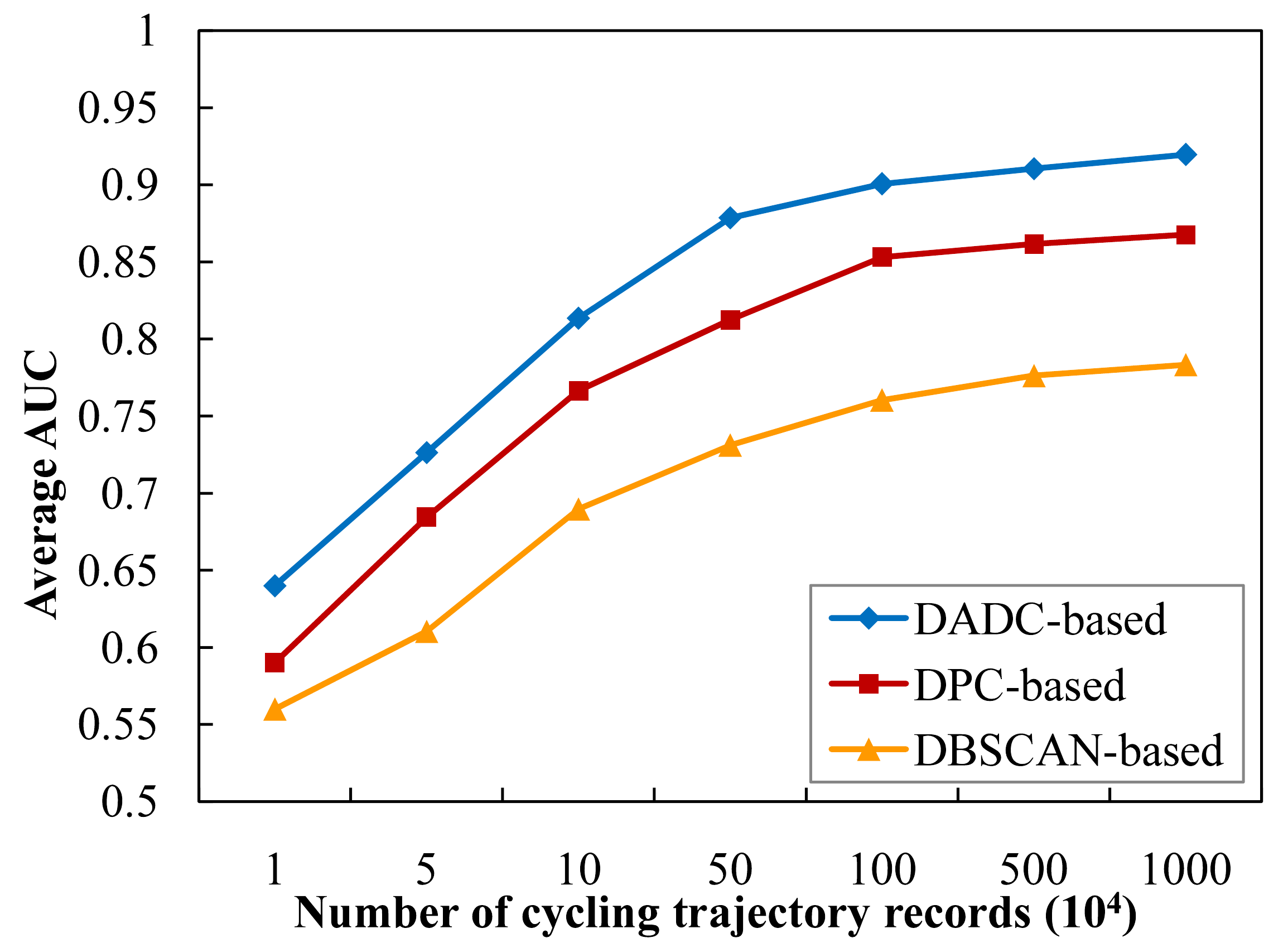}}
 \caption{Accuracy evaluation of bicycle drop-off location clustering.}
 \label{fig12}
\end{figure}

To evaluate the accuracy of DADC-based bicycle drop-off location clustering, we use two groups of DL-PBS cycling trajectory records to perform the experiments.
We matched the bicycle coordinates in the cycling trajectory records with the actual map, and then manually labeled the cluster of each data point in the two datasets.
Then, we compare the DADC method \cite{chen2019} with the Density-Peak-based clustering (DPC) \cite{ex13} and DBSCAN \cite{ester1996density} algorithms.
The average of the Area Under ROC Curve (AUC) is used as an clustering accuracy indicator by comparing the clustering results with the labels.
The experiment results are shown in Fig. \ref{fig12}.

Fig. \ref{fig12}(a) and (b) show that in all cases, the DADC-based bicycle drop-off location clustering method achieves higher AUC values than DPC and DBSCAN.
Note that cycling trajectory data has the characteristics of varying density distribution, that is, there are coexisting areas with obvious different region densities, such as dense and sparse regions (i.e., some stations have more bicycles while other stations have fewer bicycles).
DADC detects data points with dense neighbors by calculating the local density.
Based on the delta distance, DADC can efficiently identify the density peaks in each region as cluster centers.
In contrast, DPC and DBSCAN methods cannot effectively address the above issues, thereby achieving low clustering accuracy.
The average AUC value of DADC is 0.82, the average AUC value of DPC is 0.76, and that of DBSCAN is the lowest, which is 0.71.
In addition, the AUC value increases significantly with the number of cycling trajectory records.
As the number of records increases from 104 to 107, the AUC value of DADC increases from 0.63 to 0.91, the AUC value of DPC rises from 0.60 to 0.84, and that of DBSCAN only increases from 0.56 to 0.77.
Therefore, in this work, we choose the DADC method to perform bicycle drop-off location clustering.

\subsection{Evaluation of Bicycle-station Location Prediction}
\begin{figure}[!ht]
\centering
 \subfigure[Average AUC]{\includegraphics[width=2.7in]{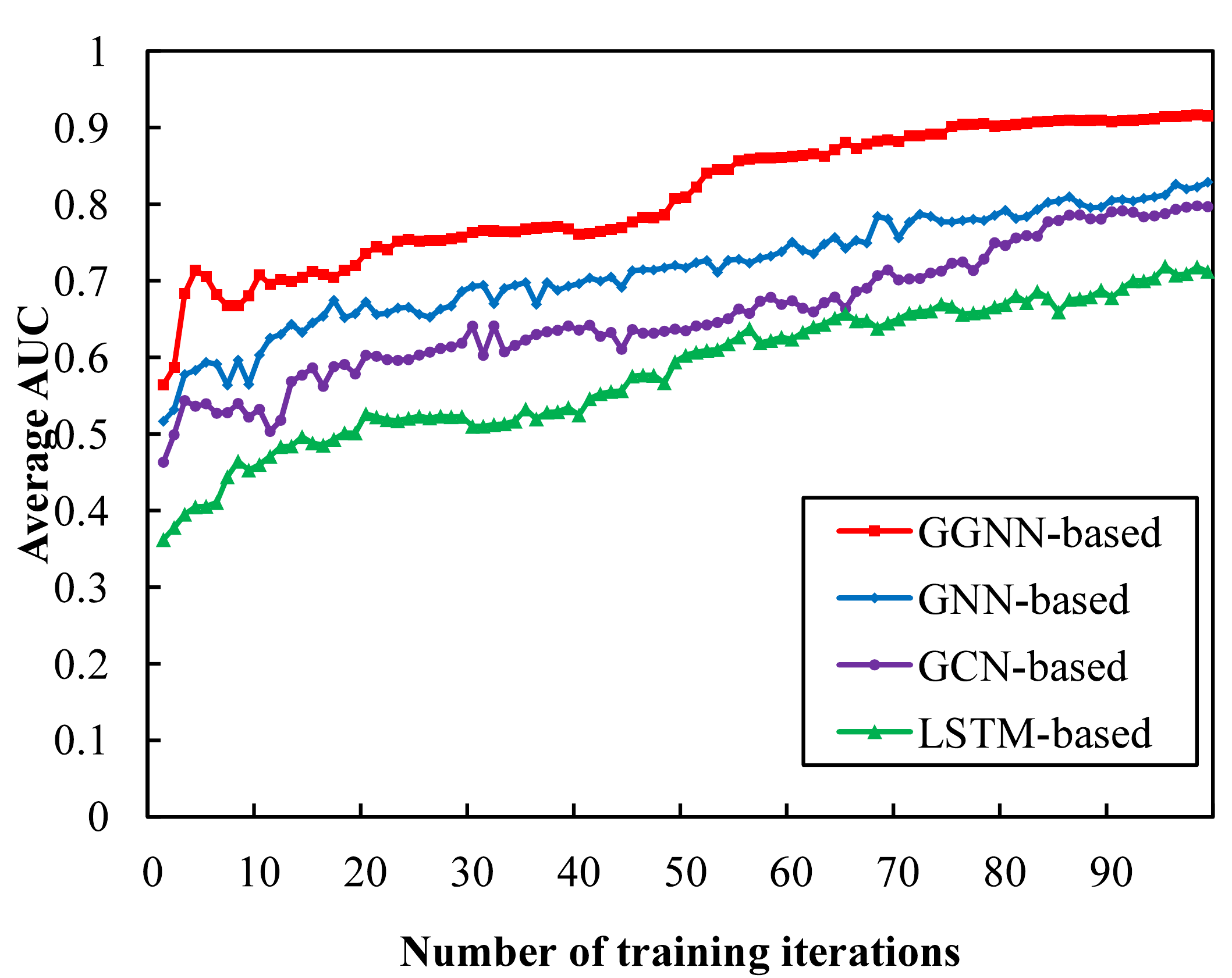}}
 \subfigure[Average RMSE]{\includegraphics[width=2.7in]{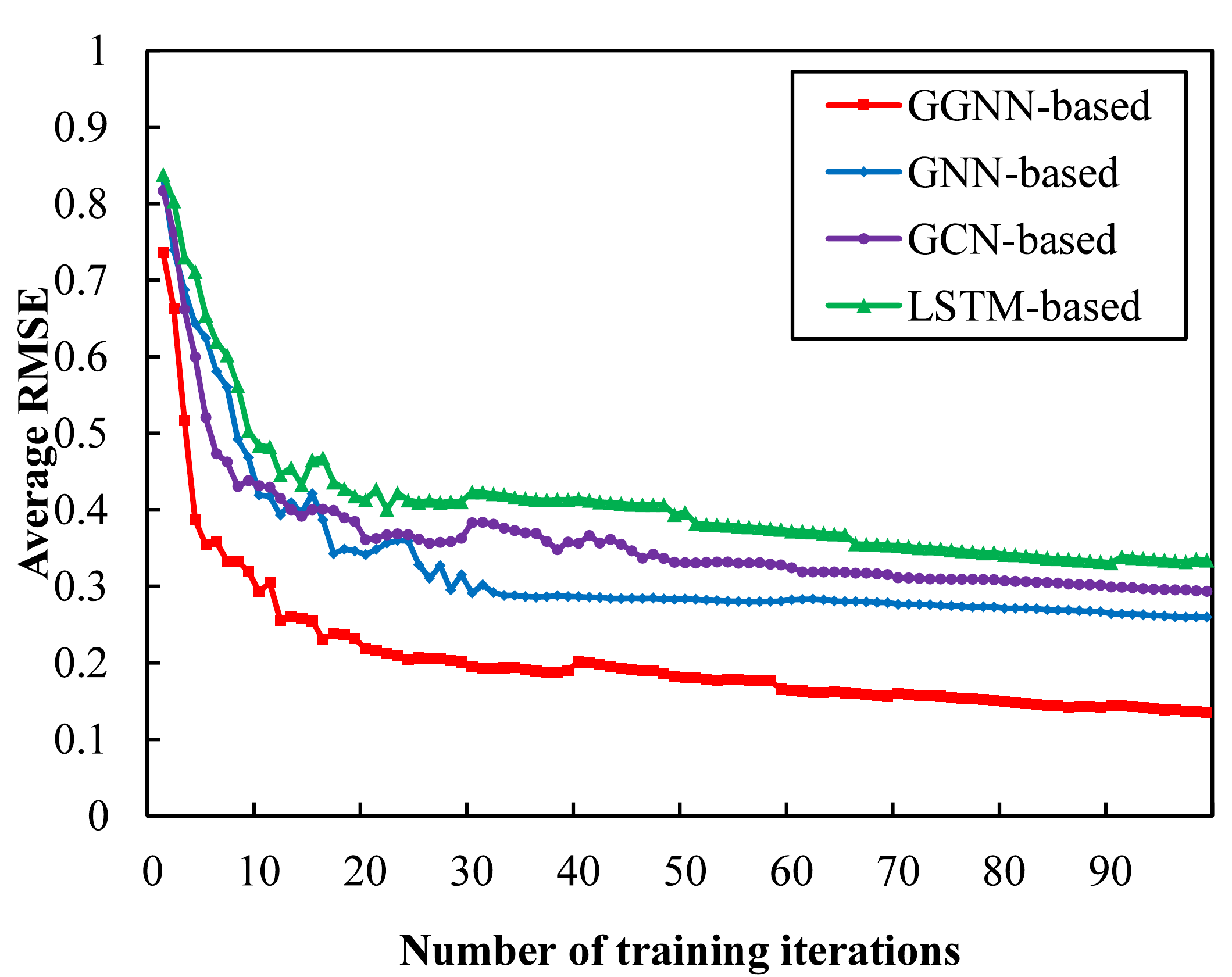}}
 \caption{Performance evaluation of bicycle station prediction.}
 \label{fig13}
\end{figure}

Based on the previous experimental results, we establish distinct graph sequence models of DL-PBS networks for different administrative regions.
To evaluate the performance of the GGNN model, we conduct experiments with these graph sequence models for bicycle station prediction by comparing GGNN \cite{gnn08}, GNN \cite{gnn05}, GCN \cite{gcn}, and LSTM \cite{lstm} models.
We discuss the performance of the comparison methods in terms of AUC and Root Mean Square Error (RMSE) \cite{hyndman2006another}, as shown in Fig. \ref{fig13}.

As shown in Fig. \ref{fig13}(a), the average AUC value of the GGNN-based bicycle station prediction method increases and reaches convergence with the number of training iterations.
Compare to other methods, GGNN can make full use of graph sequence data and capture associations between graphs in different time periods, thereby obtaining higher prediction accuracy.
In contrast, GNN and GCN methods separate the input graph at each time point, ignoring the association between these graphs.
Although the LSTM method also uses a gated module to memorize the continuous input, it lacks effective processing of the graph structure of the input.
For example, after 100 training iterations, the average AUC value of GGNN is 0.91, the average AUC value of GNN is 0.83, the average AUC value of GCN is 0.80, and the average AUC value of LSTM is 0.71.
In addition, as shown in Fig. \ref{fig13}(b), as the number of training iterations increases, the average RMSE value of GGNN decreases rapidly and converges to a minimum.
For example, after 100 iterations, the average RMSE value of GGNN is 0.13, while the average RMSE values of GGN, GCN, and LSTM are 0.26, 0.29, and 0.33, respectively.
The experimental results demonstrate that compared with the comparison methods, the GGNN model is suitable for bicycle station prediction and achieves the highest prediction accuracy.

\section{Conclusions}
\label{section6}
In this paper, we proposed a bicycle station dynamic planning (BSDP) system, which can dynamically predict DL-PBS bicycle demands and provide the optimized layout of public bicycle stations.
Firstly, by clustering large-scale historical cycling trajectory data, we established a weighted digraph model for bicycle drop-off locations.
In addition, by tracking the update of the stations in the time dimension, we further built a graph sequence model based on the bicycle drop-off digraph models.
Based on this, the GGNN model was introduced to train the graph sequence model and predict the cycling trajectory and usage requirements in the next period.
Finally, according to the requirements of public bicycle use, we provided the optimized layout of DL-PBS bicycle stations.
Experiments with actual DL-PBS datasets have verified the proposed BSDP system in terms of feasibility, accuracy, and performance.

In future work, we will focus on downstream applications of the DL-PBS network, such as public bicycle dispatching, cycling trajectory tracking, and faulty bicycle detection.

\section*{Acknowledgment}
This work is partially funded by the National Key R\&D Program of China (Grant No. 2020YFB2104000),
the National Outstanding Youth Science Program of National Natural Science Foundation of China (Grant No. 61625202),
the Program of National Natural Science Foundation of China (Grant No. 61751204),
the International (Regional) Cooperation and Exchange Program of National Natural Science Foundation of China (Grant No. 61860206011),
the Natural Science Foundation of Hunan Province (Grant No. 2020JJ5084),
and the International Postdoctoral Exchange Fellowship Program (Grant No. 20180024).
This work is also supported in part by NSF under grants III-1763325, III-1909323, and SaTC-1930941.

\bibliographystyle{abbrv}
\bibliography{sample-base}

\begin{thebibliography}{10}

\bibitem{b05}
A.~Audikana, E.~Ravalet, V.~Baranger, and V.~Kaufmann.
\newblock Implementing bikesharing systems in small cities: Evidence from the
  swiss experience.
\newblock {\em Transp. Policy}, 55:18--28, 2017.

\bibitem{b11}
R.~Cazabet, P.~Jensen, and P.~Borgnat.
\newblock Tracking the evolution of temporal patterns of usage in
  bicycle-sharing systems using nonnegative matrix factorization on multiple
  sliding windows.
\newblock {\em Int. J. Urban Sci.}, 13(4):703--712, 2017.

\bibitem{chen2018exploiting}
C.~Chen, K.~Li, S.~G. Teo, G.~Chen, X.~Zou, X.~Yang, R.~C. Vijay, J.~Feng, and
  Z.~Zeng.
\newblock Exploiting spatio-temporal correlations with multiple 3d
  convolutional neural networks for citywide vehicle flow prediction.
\newblock In {\em ICDM'18}, pages 893--898. IEEE, 2018.

\bibitem{chen2020citywide}
C.~Chen, K.~Li, S.~G. Teo, X.~Zou, K.~Li, and Z.~Zeng.
\newblock Citywide traffic flow prediction based on multiple gated
  spatio-temporal convolutional neural networks.
\newblock {\em ACM Trans. Knowl. Discov. Data}, 14(4):1--23, 2020.

\bibitem{chen2019gated}
C.~Chen, K.~Li, S.~G. Teo, X.~Zou, K.~Wang, J.~Wang, and Z.~Zeng.
\newblock Gated residual recurrent graph neural networks for traffic
  prediction.
\newblock In {\em AAAI'19}, volume~33, pages 485--492, 2019.

\bibitem{chen2019}
J.~Chen and P.~S. Yu.
\newblock A domain adaptive density clustering algorithm for data with varying
  density distribution.
\newblock {\em IEEE Trans. Knowledge Data Eng.}, 99(1):1--12, 2019.

\bibitem{b13}
C.~Deng, J.~Wang, and W.~Zheng.
\newblock Layout optimizing of public bicycle stations based on ahp in wuhan.
\newblock {\em Applied Mechanics and Materials}, 737:896--902, 2015.

\bibitem{ester1996density}
M.~Ester, H.-P. Kriegel, J.~Sander, and X.~Xu.
\newblock A density-based algorithm for discovering clusters in large spatial
  databases with noise.
\newblock In {\em KDD'96}, volume~96, pages 226--231, 1996.

\bibitem{b01}
C.~Etienne and O.~Latifa.
\newblock Model-based count series clustering for bike sharing system usage
  mining: A case study with the velib' system of paris.
\newblock {\em ACM Trans. Intell. Syst. Technol.}, 5(3):39--41, 2014.

\bibitem{b10}
N.~Gast, G.~Massonnet, D.~Reijsbergen, and M.~Tribastone.
\newblock Probabilistic forecasts of bike-sharing systems for journey planning.
\newblock In {\em CIKM'15}, pages 703--712. ACM, 2015.

\bibitem{b16}
N.~Gast, G.~Massonnet, D.~Reijsbergen, and M.~Tribastone.
\newblock Public bicycle prediction based on generalized regression neural
  network.
\newblock In {\em IOV'15}, pages 363--373, Sichuan, China, 2015. Springer.

\bibitem{lstm}
S.~Hochreiter and J.~Schmidhuber.
\newblock Long short-term memory.
\newblock {\em Neural Comput.}, 9(8):1735--1780, 1997.

\bibitem{b04}
A.~Hofleitner, R.~Herring, P.~Abbeel, and A.~Bayen.
\newblock Learning the dynamics of arterial traffic from probe data using a
  dynamic bayesian network.
\newblock {\em IEEE Trans. Intell. Transp. Syst.}, 13(4):1679--1693, 2012.

\bibitem{hyndman2006another}
R.~J. Hyndman and A.~B. Koehler.
\newblock Another look at measures of forecast accuracy.
\newblock {\em Int. J. Forecast.}, 22(4):679--688, 2006.

\bibitem{jia2016location}
Y.~Jia, Y.~Wang, X.~Jin, and X.~Cheng.
\newblock Location prediction: A temporal-spatial bayesian model.
\newblock {\em ACM Trans. Intell. Syst. Technol.}, 7(3):1--25, 2016.

\bibitem{b14}
Z.~Jiang, M.~Evans, D.~Oliver, and S.~Shekhar.
\newblock Identifying k primary corridors from urban bicycle gps trajectories
  on a road network.
\newblock {\em Inf. Syst.}, 57:142--159, 2016.

\bibitem{gnn03}
M.~Khodayar and J.~Wang.
\newblock Spatio-temporal graph deep neural network for short-term wind speed
  forecasting.
\newblock {\em IEEE Trans. Sustain. Energy}, 10(2):670--681, 2019.

\bibitem{gnn01}
D.~Krleza and K.~Fertalj.
\newblock Graph matching using hierarchical fuzzy graph neural networks.
\newblock {\em IEEE Trans. Fuzzy Syst.}, 25(4):892--904, 2017.

\bibitem{b06}
K.~Labadi, T.~Benarbia, J.-P. Barbot, S.~Hamaci, and A.~Omari.
\newblock Stochastic petri net modeling, simulation and analysis of public
  bicycle sharing systems.
\newblock {\em IEEE Trans. Autom. Sci. Eng.}, 12(4):1380--1395, 2015.

\bibitem{gnn04}
R.~Levie, F.~Monti, X.~Bresson, and M.~M. Bronstein.
\newblock Cayleynets: Graph convolutional neural networks with complex rational
  spectral filters.
\newblock {\em IEEE Trans. Signal Process.}, 67(1):97--109, 2019.

\bibitem{gnn08}
Y.~Li, D.~Tarlow, M.~Brockschmidt, and R.~Zemel.
\newblock Gated graph sequence neural networks.
\newblock {\em arXiv e-prints}, June 2017.

\bibitem{li2015traffic}
Y.~Li, Y.~Zheng, H.~Zhang, and L.~Chen.
\newblock Traffic prediction in a bike-sharing system.
\newblock In {\em SIGSPATIAL'15}, pages 1--10, Seattle, Washington, USA, 2015.
  ACM.

\bibitem{b15}
L.~Lin, Z.~He, and S.~Peeta.
\newblock Predicting station-level hourly demand in a large-scale bike-sharing
  network: A graph convolutional neural network approach.
\newblock {\em Transp. Res. Pt. C-Emerg. Technol.}, 97:258--276, 2018.

\bibitem{b03}
Y.~Ma, X.~Qin, J.~Xu, and W.~Wang.
\newblock A hierarchical public bicycle dispatching policy for dynamic demand.
\newblock In {\em SOLI'16}, pages 162--167, Beijing, China, 2016. IEEE.

\bibitem{gnn06}
A.~Micheli.
\newblock Neural network for graphs: A contextual constructive approach.
\newblock {\em IEEE Trans. Neural Netw.}, 20(3):498--511, 2009.

\bibitem{pan2019deep}
L.~Pan, Q.~Cai, Z.~Fang, P.~Tang, and L.~Huang.
\newblock A deep reinforcement learning framework for rebalancing dockless bike
  sharing systems.
\newblock In {\em AAAI'19}, volume~33, pages 1393--1400, 2019.

\bibitem{ex13}
A.~Rodriguez and A.~Laio.
\newblock Clustering by fast search and find of density peaks.
\newblock {\em Science}, 344(6191):1492--1496, 2014.

\bibitem{gnn05}
F.~Scarselli, M.~Gori, A.~C. Tsoi, M.~Hagenbuchner, and G.~Monfardini.
\newblock The graph neural network model.
\newblock {\em IEEE Trans. Neural Netw.}, 20(1):61--80, 2009.

\bibitem{tii01}
Z.~Xiao, H.~B. Lim, and L.~Ponnambalam.
\newblock Participatory sensing for smart cities: A case study on transport
  trip quality measurement.
\newblock {\em IEEE Trans. Ind. Inform.}, 13(2):759--770, 2017.

\bibitem{ying2014mining}
J.~J.-C. Ying, W.-C. Lee, and V.~S. Tseng.
\newblock Mining geographic-temporal-semantic patterns in trajectories for
  location prediction.
\newblock {\em ACM Trans. Intell. Syst. Technol.}, 5(1):1--33, 2014.

\bibitem{gcn}
R.~Ying, R.~He, K.~Chen, P.~Eksombatchai, W.~L. Hamilton, and J.~Leskovec.
\newblock Graph convolutional neural networks for web-scale recommender
  systems.
\newblock In {\em KDD}, pages 974--983. ACM, 2018.

\bibitem{zhang2017bi}
L.~Zhang, K.~Li, C.~Li, and K.~Li.
\newblock Bi-objective workflow scheduling of the energy consumption and
  reliability in heterogeneous computing systems.
\newblock {\em Inf. Sci.}, 379:241--256, 2017.

\bibitem{zhang2017contention}
L.~Zhang, K.~Li, W.~Zheng, and K.~Li.
\newblock Contention-aware reliability efficient scheduling on heterogeneous
  computing systems.
\newblock {\em IEEE Trans. Sustain. Comput.}, 3(3):182--194, 2017.

\end{thebibliography}

\end{document}